\documentclass[review]{elsarticle}

\usepackage[hyphens]{url}
\usepackage{hyperref}
\hypersetup{breaklinks=true}

\usepackage[utf8]{inputenc}

\usepackage{graphicx}
\graphicspath{ {resources/} }
\usepackage{tkz-graph}
\usetikzlibrary{arrows, automata}
\usepackage{amsmath}
\allowdisplaybreaks[1] 
\usepackage{mathtools}
\usepackage[ruled,vlined,linesnumbered]{algorithm2e}
\usepackage{algorithmic}
\usepackage{float} 
\usepackage{paralist} 

\usepackage[hang, small,labelfont=bf,up,textfont=it,up]{caption}
\usepackage[sc]{mathpazo} 
\usepackage[T1]{fontenc} 
\usepackage{color}
\usepackage{xcolor}

\usepackage{lineno,hyperref}
\modulolinenumbers[5]

\usepackage{etoolbox}
\makeatletter
\patchcmd{\ps@pprintTitle}
  {Preprint submitted to}
  {}
  {}{}
\makeatother

\journal{Technical report}

\begin{document}

\begin{frontmatter}

\title{On the Min-cost Traveling Salesman Problem with Drone}

\author{Quang Minh Ha}
\address{quang.ha@uclouvain.be}
\address{ICTEAM, Université catholique de Louvain, Belgium}

\author{Yves Deville}
\address{yves.deville@uclouvain.be}
\address{ICTEAM, Université catholique de Louvain, Belgium}

\author{Quang Dung Pham}
\address{dungpq@soict.hust.edu.vn}
\address{SoICT, Hanoi University of Science and Technology, Vietnam}

\author{Minh Hoàng Hà\corref{mycorrespondingauthor}}
\cortext[mycorrespondingauthor]{Corresponding author}
\address{minhhoang.ha@vnu.edu.vn}
\address{University of Engineering and Technology, Vietnam National University, Vietnam}

\begin{abstract}
Over the past few years, unmanned aerial vehicles (UAV), also known as drones, have been adopted as part of a new logistic method in the commercial sector called "last-mile delivery". In this novel approach, they are deployed alongside trucks to deliver goods to customers to improve the quality of service and reduce the transportation cost. This approach gives rise to a new variant of the traveling salesman problem (TSP), called TSP with drone (TSP-D). A variant of this problem that aims to minimize the time at which truck and drone finish the service (or, in other words, to maximize the quality of service) was studied in the work of Murray and Chu (2015). In contrast, this paper considers a new variant of TSP-D in which the objective is to minimize operational costs including total transportation cost and one created by waste time a vehicle has to wait for the other. The problem is first formulated mathematically. Then, two algorithms are proposed for the solution. The first algorithm (TSP-LS) was adapted from the approach proposed by Murray and Chu (2015), in which an optimal TSP solution is converted to a feasible TSP-D solution by local searches. The second algorithm, a Greedy Randomized Adaptive Search Procedure (GRASP), is based on a new split procedure that optimally splits any TSP tour into a TSP-D solution. After a TSP-D solution has been generated, it is then improved through local search operators. Numerical results obtained on various instances of both objective functions with different sizes and characteristics are presented. The results show that GRASP outperforms TSP-LS in terms of solution quality under an acceptable running time.
\end{abstract}


\end{frontmatter}

\linenumbers

\section{Introduction}
\label{section:introduction}

Companies always tend to look for the most cost-efficient methods to distribute goods across logistic networks \cite{rizzoli2007ant}. Traditionally, trucks have been used to handle these tasks and the corresponding transportation problem is modelled as a traveling salesman problem (TSP). However, a new distribution method has recently arisen in which small unmanned aerial vehicles (UAV), also known as drones, are deployed to support parcel delivery. On the one hand, there are four advantages of using a drone for delivery: (1) it can be operated without a human pilot, (2) it avoids the congestion of traditional road networks by flying over them, (3) it is faster than trucks, and (4) it has much lower transportation costs per kilometre \cite{marcuswohlsen2013}. On the other hand, because the drones are powered by batteries, their flight distance and lifting power are limited, meaning they are restricted in both maximum travel distance and parcel size. In contrast, a truck has the advantage of long range travel capability. It can carry large and heavy cargo with a diversity of size, but it is also heavy, slow and has much higher transportation cost.

Consequently, the advantages of truck offset the disadvantages of drones and — similarly — the advantages of drones offset the disadvantages of trucks. These complementary capabilities are the foundation of a novel method named "last mile delivery with drone" \cite{stevebanker2013}, in which the truck transports the drone close to the customer locations, allowing the drone to service customers while remaining within its flight range, effectively increasing the usability and making the schedule more flexible for both drones and trucks. Specifically, a truck departs from the depot carrying the drone and all the customer parcels. As the truck makes deliveries, the drone is launched from the truck to service a nearby customer with a parcel. While the drone is in service, the truck continues its route to further customer locations. The drone then returns to the truck at a location different from its launch point.

From the application perspective, a number of remarkable events have occurred since 2013, when Amazon CEO Jeff Bezos first announced Amazon's plans for drone delivery \cite{cbsnews2013}, termed "a big surprise." Recently, Google has been awarded a patent that outlines its drone delivery method \cite{mikemurphy2016}. In detail, rather than trying to land, the drone will fly above the target, slowly lowering packages on a tether. More interestingly, it will be able to communicate with humans using voice messages during the delivery process. Google initiated this important drone delivery project, called Wing, in 2014, and it is expected to launch in 2017 \cite{michaelgrothaus2016}. A similar Amazon project called Amazon Prime Air ambitiously plans to deliver packages by drone within 30 minutes \cite{davidpogue2016}. Other companies worldwide have also been testing delivery services using drones. In April 2016, Australia Post successfully tested drones for delivering small packages. That project is reportedly headed towards a full customer trial later this year \cite{anthonycuthbertson2016}. In May 2016, a Japanese company---Rakuten--- launched a service named "Sora Kaku" that "delivers golf equipment, snacks, beverages and other items to players at pickup points on the golf course" \cite{ctvnews2016}. In medical applications, Matternet, a California-based startup, has been testing drone deliveries of medical supplies and specimens (such as blood samples) in many countries since 2011. According to their CEO: it is "much more cost-, energy- and time-efficient to send [a blood sample] via drone, rather than send it in a two-ton car down the highway with a person inside to bring it to a different lab for testing," \cite{sallyfrench2015}. Additionally, a Silicon Valley start-up named Zipline International began using drones to deliver medicine in Rwanda starting in July, 2016 \cite{amartoor2016}.

We are aware of several publications in the literature that have investigated the routing problem related to the truck-drone combination for delivery. Murray and Chu~\cite{murray2015flying} introduced the problem, calling it the "Flying Sidekick Traveling Salesman Problem" (FSTSP). A mixed integer liner programming (MILP) formulation and a heuristic are proposed. Basically, their heuristic is based on a "Truck First, Drone Second" idea, in which they first construct a route for the truck by solving a TSP problem and, then, repeatedly run a relocation procedure to reduce the objective value. In detail, the relocation procedure iteratively checks each node from the TSP tour and tries to consider whether it is suitable for use as a drone node. The change is applied immediately when this is true, and the current node is never checked again. Otherwise, the node is relocated to other positions in an effort to improving the objective value. The relocation procedure for TSP-D is designed in a "best improvement" fashion; it evaluates all the possible moves and executes the best one. The proposed methods are tested on only small-sized instances with up to 10 customers.

Agatz et al.~\cite{agatz2016optimization}, study a slightly different problem---called the "Traveling Salesman Problem with Drone" (TSP-D), in which the drone has to follow the same road network as the truck. Moreover, in TSP-D, the drone may be launched and return to the same location, while this is forbidden in the FSTSP. This problem is also modelled as a MILP formulation and solved by a "Truck First, Drone Second" heuristic in which drone route construction is based on either local search or dynamic programming. More recently, Ponza~\cite{ponza2016optimization} extended the work of Murray and Chu~\cite{murray2015flying} in his master's thesis to solve the FSTSP, proposing an enhancement to the MILP model and solving the problem by a heuristic method based on Simulated Annealing.

Additionally, Wang et al. \cite{wang2016vehicle}, in a recent research, introduced a more general problem that copes with multiple trucks and drones with the goal of minimizing the completion time. The authors named the problem "The vehicle routing problem with drone" (VRP-D) and conducted the analysis on several worst-case scenarios, from which they propose bounds on the best possible savings in time when using drones and trucks instead of trucks alone.

All the works mentioned above aim to minimize the time at which the truck and the drone complete the route and return to the depot, which can improve the quality of service \cite{nozick2001inventory}. However, in every logistics activities, operational costs also play an important role in the overall business cost (see \cite{dawnrusselljohnj2014} and \cite{adamrobinson2014}). Hence, minimizing these costs by using a more cost-efficient approach is a vital objective of every company involved in transport and logistics activities. Recently, an objective function that minimizes the transportation cost was studied by Mathew et al.~\cite{mathew2015planning} for a related problem called the Heterogeneous Delivery Problem (HDP). However, unlike in \cite{murray2015flying}, \cite{agatz2015optimization} and \cite{ponza2016optimization}, the problem is modelled on a directed physical street network where a truck cannot achieve direct delivery to the customer. Instead, from the endpoint of an arc, the truck can launch a drone that will service the customers. In this way, the problem can be favourably transformed to a Generalized Traveling Salesman Problem (GTSP) \cite{gutin2006traveling}. The authors use the Nood-Bean Transformation available in Matlab to reduce a GTSP to a TSP, which is then solved by a heuristic proposed in the study. To the best of our knowledge, the min-cost objective function has not been studied for TSP-D when the problem is defined in a more realistic way---similarly to \cite{murray2015flying}, \cite{agatz2015optimization}, and \cite{ponza2016optimization}. Consequently, this gap in the literature provides a strong motivation for studying TSP-D with the min-cost objective function.

This paper studies a new variant of TSP-D following the hypotheses of the FSTSP proposed in the work of \cite{murray2015flying}. In FSTSP, the objective is to minimize the delivery completion time, or in other word the time coming back to the depot, of both truck and drone. In the new variant that we call min-cost TSP-D, the objective is to minimize the total operational cost of the system including two distinguished parts. The first part is the transportation cost of truck and drone while the second part relates to the waste time a vehicle has to wait for the other whenever drone is launched. In the following, we denote the FSTSP as min-time TSP-D to avoid confusion. 

In this paper, we propose a MILP model and two heuristics to solve the min-cost TSP-D: a Greedy Randomized Adaptive Search Procedure (GRASP) and a heuristic adapted from the work of \cite{murray2015flying} called TSP-LS. In detail, the contributions of this paper are as follows:
\begin{itemize}
\item[-] We introduce a new variant of TSP-D called min-cost TSP-D, in which the objective is to minimize the operational costs.
\item[-] We propose a model together with a MILP formulation for the problem which is an extended version of the model proposed in \cite{murray2015flying} for min-time TSP-D.
\item[-] We develop two heuristics for min-cost TSP-D: TSP-LS and GRASP.  which contain a new split procedure and local search operators. We also adapt our solution methods to solve the min-time problem studied in \cite{murray2015flying}.
\item[-] We introduce various sets of instances with different numbers of customers and a variety of options to test the problem. 
\item[-] We conduct various experiments to test our heuristics on the min-cost as well as min-time problems. We also compare solutions of both objectives. The computational results show that GRASP outperforms TSP-LS in terms of quality of solutions with an acceptable running time. TSP-LS delivers solution of lower quality, but very quickly. 
\end{itemize}

This article is structured as follows: Section \ref{section:introduction} provides the introduction. Section \ref{section:problem-definition} describes the problem and the model. The MILP formulation is introduced in Section \ref{section:mip}. We describe our two heuristics in Sections \ref{section:grasp} and \ref{section:TSP-LS}. Section \ref{section:experiments} presents the experiments, including instance generations and settings. We discuss the computational results in Section \ref{section:results}. Finally, Section \ref{section:conclusion} concludes the work and provides suggestions for future research.

\section{Problem definition}
\label{section:problem-definition}

In this section, we provide a description of the problem and discuss a model for the min-cost TSP-D in a step-by-step manner. Here, we consider a list of customers to whom a truck and a drone will deliver parcels. To make a delivery, the drone is launched from the truck and later rejoins the truck at another location. Each customer is visited only once and is serviced by either the truck or the drone. Both vehicles must start from and return to the depot. When a customer is serviced by the truck, this is called a \textbf{truck delivery}, while when a customer is serviced by the drone, this is called a \textbf{drone delivery}. This is represented as a 3-tuple $\langle i, j, k\rangle$, where $i$ is a launch node, $j$ is a drone node (a customer who will be serviced by the drone), and $k$ is a rendezvous node, as listed below:
\begin{itemize}
\item Node $i$ is a launch node at which the truck launches the drone. The launching operation must be carried out at a customer location or the depot. The time required to launch the drone is denoted as $s_L$.
\item Node $j$ is a node serviced by the drone, called a "drone node". We also note that not every node in the graph is a drone node. Because some customers might demand delivery a product with size and weight larger than the capacity of the drone.
\item Node $k$ is a customer location where the drone rejoins the truck. At node $k$, the two vehicles meet again; therefore, we call it "rendezvous node". While waiting for the drone to return from a delivery, the truck can make other truck deliveries. The time required to retrieve the drone and prepare for the next drone delivery is denoted as $s_R$. Moreover, the two vehicles can wait for each other at the rendezvous point.
\end{itemize}
Moreover, the drone has an "endurance", which can be measured as the maximum time the drone can operate without recharging. A tuple $\langle i, j, k\rangle$ is called feasible if the drone has sufficient power to launch from $i$, deliver to $j$ and rejoin the truck at $k$. The drone can be launched from the depot but must subsequently rejoin the truck at a customer location. Finally, the drone's last rendezvous with the truck can occur at the depot.

When not actively involved in a \textbf{drone delivery}, the drone is carried by the truck. We also assume that the drone is in constant flight when waiting for the truck. Furthermore, the truck and the drone have their own transportation costs per unit of distance. In practice, the drone's cost is much lower than the truck's cost because it weighs much less than the truck, hence, consuming much less energy. In addition, it is not run by gasoline but by batteries. We also assume that the truck provides new fresh batteries for the drone (or recharges its batteries completely) before each drone delivery begins. When a vehicle has to wait for each other, a penalty is created and added to the transportation cost to form the total operational cost of the system. The waiting costs of truck and drone are calculated by: 
\begin{equation}
	\text{waiting cost}_{truck} = \alpha \times \text{waiting time} \nonumber
\end{equation}
\begin{equation}
	\text{waiting cost}_{drone} = \beta \times \text{waiting time} \nonumber
\end{equation}
where $\alpha$ and $\beta$ are the waiting fees of truck and drone per unit of time, respectively.

\begin{figure}[H] 
\centering
(a) Optimal TSP tour \\
\includegraphics[scale=0.4]{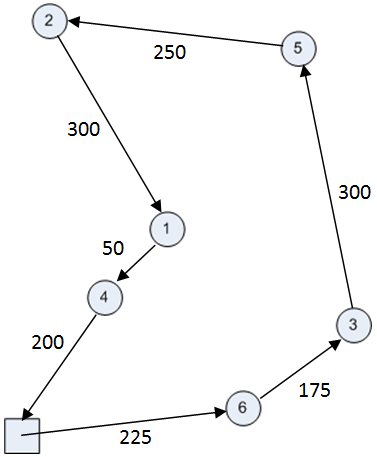} \\
(b) Optimal min-cost TSP-D tour \\
\includegraphics[scale=0.4]{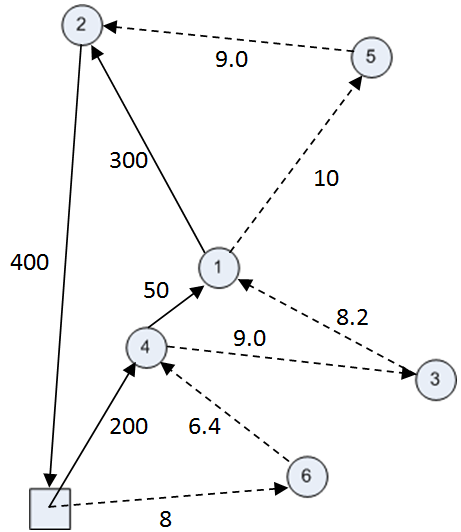}
\caption{Optimal solution: TSP vs. min-cost TSP-D. \\ TSP Objective = 1500, min-cost TSP-D Objective = 1000.82. The solid arcs are truck's path. The dash arcs are drone's path}\label{figure:tsp-tspd}
\end{figure}

The objective of the min-cost TSP-D is to minimize the total operational cost of the system which includes the travel cost of truck and drone as well as their waiting costs. Because the problem reduces to a TSP when the drone's endurance is null, it is NP-Hard. Examples of TSP and min-cost TSP-D optimal solutions on the same instance in which the unitary transportation cost of the truck is 25 times more expensive than that of the drone are shown in Figure \ref{figure:tsp-tspd}.

We now develop the model for the problem. We first define basic notations relating to the graph, sequence and subsequence. Then, we formally define drone delivery and the solution representation as well as the associated constraints and objective.

\subsection{The min-cost TSP-D problem}
The min-cost TSP-D is defined on a graph $G = (V, A)$, $V = \{0, 1, \ldots, n, n + 1\}$, where $0$ and $n + 1$ both represent the same depot but are duplicated to represent the starting and returning points. The set of customers is $N = \{1, \ldots, n\}$. Let $V_D \subseteq N$ denote the set of customers that can be served by drone. Let $d_{ij}$ and $d'_{ij}$ be the distances from node $i$ to node $j$ travelled by the truck and the drone, respectively. We also denote $\tau_{ij}$, $\tau'_{ij}$ the travel time of truck and drone from $i$ to $j$. Furthermore, $C_1$ and $C_2$ are the transportation costs of the truck and drone, respectively, per unit of distance. 

Given a sequence $s = \langle s_1, s_2, \ldots, s_t \rangle$, where $s_i \in V, i = 1 \ldots t$, we denote the following:
\begin{itemize}
\item[-] $V(s) \subseteq V$ the list of nodes of $s$ 
\item[-] $pos(i, s)$ the position of node $i \in V$ in $s$ 
\item[-] $next_s(i)$, $prev_s(i)$ the next node and previous node of $i$ in $s$
\item[-] $first(s)$, $last(s)$ the first node and last node of $s$
\item[-] $s[i]$ the $i^{th}$ node in $s$
\item[-] $size(s)$ the number of nodes in $s$ 
\item[-] $sub(i, j, s)$, where $\forall i, j \in s, pos(i,s) < pos(j,s)$, the subsequence of $s$ from node $i$ to node $j$
\item[-] $A(s) = \{(i, next_s(i)) | i \in V(s) \setminus last(s)\}$ the set of arcs in $s$
\item[-] $d_{i \rightarrow k}$ the distance traveled by truck from $i$ to $k$ in the truck tour
\item[-] $t_{i \rightarrow k}$ the time traveled by truck from $i$ to $k$ in the truck tour
\item[-] $t'_{ijk}$ the time traveled by drone from $i$ to $j$ to $k$ in a drone delivery
\item[-] $d^{-j}_{i \rightarrow k}$ the distance traveled by truck from $i$ to $k$ in the truck tour with $j$ removed ($j$ is between $i$ and $k$)
\end{itemize}

As mentioned above, we define a \textbf{drone delivery} as a 3-tuple $\langle i, j, k \rangle : i, j, k \in V, i \neq j, j \neq k, k \neq i, \tau'_{ij} + \tau'_{jk} \leq \epsilon$, where $\epsilon$ is a constant denoting the drone's endurance. We also denote $\mathbb{P}$ as the set of all possible drone deliveries on the graph $G = (V,A)$ that satisfies the endurance constraint as follows:
\begin{align}
\mathbb{P} = \{ \langle i, j, k \rangle : i, k \in V, j \in V_D, i \neq j, j \neq k, k \neq i, \tau'_{ij} + \tau'_{jk} \leq \epsilon \}\nonumber.
\end{align}

\subsection{Solution representation}
A \textbf{min-cost TSP-D solution}, denoted as $sol$, is represented by two components:
\begin{itemize}
\item[-] A truck tour, denoted as \textbf{$TD$}, is a sequence $\langle e_0, e_1, \ldots, e_k \rangle$, where $e_0 = e_k = 0, e_i \in V$, $e_i \neq e_j$ and $i \neq j$.
\item[-] A set of drone deliveries \textbf{$DD$} such that $DD \subseteq \mathbb{P}$, 
\end{itemize}
which can also be written as
\begin{align}
sol = (TD, DD). \nonumber
\end{align}

\subsection{Constraints}
A solution ($TD, DD$) of the min-cost TSP-D must satisfy the following constraints:
\begin{itemize}
\item[(A)] Each customer must be serviced by either the truck or the drone:
\begin{align}
\forall e \in N: e \in TD \text{ or } \exists \langle i, e, k \rangle \in DD
\nonumber.
\end{align}
By definition, during a truck tour, a customer cannot be visited twice by the truck. The above constraint does not prevent a customer from being serviced by both the truck and the drone nor from being serviced twice by the drone.
\item[(B)] A customer is never serviced twice by the drone:
\begin{align}
\forall \langle i, j, k \rangle, \langle i', j', k' \rangle \in DD : j \neq j'
\nonumber.
\end{align}
\item[(C)] Drone deliveries must be compatible with the truck tour:
\begin{align}
\forall \langle i, j, k \rangle \in DD : j \notin TD, i \in TD, k \in TD, pos(i, TD) < pos(k, TD)
\nonumber.
\end{align}
This constraint implies that a customer cannot be serviced by both the truck and drone.
\item[(D)] No interference between drone deliveries:
\begin{align}
\forall \langle i, \cdot, k \rangle \in DD, \forall e \in sub(i, k, TD), \forall \langle i', j', k' \rangle \in DD: e \neq i'
\nonumber.
\end{align}
The above constraint means that when the drone is launched from the truck for a drone delivery, it cannot be relaunched before the rendezvous from that delivery. As a consequence, we cannot have any other rendezvous during that period either.
\end{itemize}

\subsection{Objective}
Regarding the costs, the following notations are used:
\begin{itemize}
\item[-] $cost(i, j, k) = C_2 (d'_{ij} + d'_{jk})$, where $\langle i, j, k \rangle \in \mathbb{P}$ [cost of drone delivery $\langle i, j, k \rangle$]
\item[-] $cost^T_W(i, j, k) = \alpha \times max(0, (t_{i \rightarrow k} - t'_{ijk}))$, where $\langle i, j, k \rangle \in \mathbb{P}$ [waiting cost of truck at $k$]
\item[-] $cost^D_W(i, j, k) = \beta \times max(0, (t'_{ijk} - t_{i \rightarrow k}))$, where $\langle i, j, k \rangle \in \mathbb{P}$ [waiting cost of drone at $k$]
\item[-] $cost(TD) = \mathlarger{\sum_{(i,j) \in A(TD)} C_1.d_{ij}}$ [cost of a truck tour $TD$]
\item[-] $cost(DD) = \mathlarger{\sum_{\langle i, j, k \rangle \in DD}cost(i, j, k)}$ [total cost of all drone deliveries in $DD$]
\item[-] $cost_W(DD) = \mathlarger{\sum_{\langle i, j, k \rangle \in DD}cost^T_W(i, j, k) + cost^D_W(i, j, k)}$ [total waiting cost]
\item[-] $cost(TD, DD) = cost(TD) + cost(DD) + cost_W(DD)$ [cost of a solution] 
\item[-] $cost(sub(i, k, s))$ the total cost for both truck and drone and their waiting cost (if any) in a subsequence $s \in TD$.
\end{itemize}

\textbf{The objective} is to minimize the total operational cost:
\begin{align}
\min \text{ } cost(TD, DD).
\nonumber
\end{align}

\section{Mixed Integer Linear Programming Formulation}
\label{section:mip}

The min-cost TSP-D defined in the previous section is represented here in a MILP formulation. This formulation is an extension from the one proposed by \cite{murray2015flying}. We extend it by proposing constraints where waiting time is captured in order to calculate the waiting cost of two vehicles. We first define two subsets of $V$, $V_L = \{ 0, 1, \ldots, n\}$ and $V_R = \{1, 2, \ldots, n + 1\}$ to distinguish the nodes that from where the drone can be launched from and the one it returns to.

\textbf{Variables}

Let $x_{ij} \in \{0, 1\}$ equal one if the truck goes from node $i$ to node $j$ with $i \in V_L$ and $j \in V_R$, $i \neq j$. Let $y_{ijk} \in \{0, 1\}$ equal one if $\langle i, j , k \rangle$ is a \textbf{drone delivery}. We can denote $p_{ij} \in \{0, 1\}$ as equalling one if node $i \in N$ is visited before node $j \in N$, $j \neq i$, in the truck's path. We also set $p_{0j} = 1$ for all $j \in N$ to indicate that the truck always starts the tour from the depot. As in standard TSP subtour elimination constraints, we denote $0 \leq u_i \leq n + 1$ as the position of the node $i$, $i \in V$ in the truck's path. 

To handle the waiting time, let $t_i \geq 0, t'_i \geq 0, i \in V_R$ denote the arrival time of truck and drone at node $i$, $r_i \geq 0, r'_i \geq 0, i \in V_R$ the leaving time of truck and drone at node $i$. We also denote $w_i \geq 0, w'_i \geq 0, i \in V_R$ the waiting time of truck and drone at node $i$ respectively. Finally, we have $t_0 = 0, t'_0 = 0, r_0 = 0, r'_0 = 0$ the earliest time of truck and drone starting from depot $0$ and $w_0 = 0, w'_0 = 0$ the waiting time at the starting depot.

\textbf{The MILP formulation} is as follows:


\begin{align}
\text{Min } & \mathlarger{C_1\sum\limits_{\substack{i \in V_L}}\sum\limits_{\substack{j \in V_R \\ i \neq j}} d_{ij} x_{ij}} + \mathlarger{C_2\sum\limits_{\substack{i \in V_L}}\sum\limits_{\substack{j \in N \\ i \neq j}}\sum\limits_{\substack{k \in V_R \\ \langle i,j,k \rangle \in \mathbb{P}}} (d'_{ij} + d'_{jk}) y_{ijk}} \nonumber \\ \label{constraint:1}
& \quad\quad + \mathlarger{ \alpha\sum\limits_{\substack{i \in V}}w_i + \beta\sum\limits_{\substack{i \in V}}w'_i} 
\end{align}


\begin{align}
& \mathlarger{\sum\limits_{\substack{i \in V_L \\ i \neq j}}x_{ij} + \sum\limits_{\substack{i \in V_L \\ i \neq j}}\sum\limits_{\substack{k \in V_R \\ \langle i, j, k \rangle \in \mathbb{P}}}y_{ijk} = 1 \quad \forall j \in N} \label{constraint:2} \\
& \mathlarger{\sum\limits_{j \in V_R}x_{0j} = 1} \label{constraint:3} \\
& \mathlarger{\sum\limits_{i \in V_L}x_{i, n + 1} = 1} \label{constraint:4} \\
& \mathlarger{u_i - u_j + 1 \leq (n + 2)(1 - x_{ij})} \quad \forall i \in V_L, j \in \{V_R: i \neq j\} \label{constraint:5} \\
& \mathlarger{\sum\limits_{\substack{i \in V_L \\ i \neq j}}x_{ij} = \sum\limits_{\substack{k \in V_R \\ k \neq j}}x_{jk} \quad \forall j \in N} \label{constraint:6}
\end{align}


\begin{align}
& \mathlarger{2 y_{ijk} \leq \sum\limits_{\substack{h \in V_L \\ h \neq i}}x_{hi} + \sum\limits_{\substack{l \in N \\ l \neq k}}x_{lk}} \label{constraint:7} \\
& \quad \forall i \in N, j \in \{N: i \neq j\}, k \in \{V_R: \langle i, j, k \rangle \in \mathbb{P} \} \nonumber \\ 
& \mathlarger{y_{0jk} \leq \sum\limits_{\substack{h \in V_L \\ h \neq k \\ h \neq j}}x_{hk} \quad j \in N, k \in \{V_R: \langle 0, j, k \rangle \in \mathbb{P}\}} \label{constraint:8} \\
& \mathlarger{u_k - u_i \geq 1 - (n + 2)(1 - \sum\limits_{\substack{j \in N \\ \langle i, j ,k \rangle \in \mathbb{P}}}y_{ijk})} \label{constraint:9} \\
& \quad \forall i \in V_L, k \in \{V_R: k \neq i\} \nonumber 
\end{align}


\begin{align}
& \mathlarger{\sum\limits_{\substack{j \in N \\ j \neq i}}\sum\limits_{\substack{k \in V_R \\ \langle i, j, k \rangle \in \mathbb{P}}}y_{ijk} \leq 1 \quad \forall i \in V_L} \label{constraint:10} \\
& \mathlarger{\sum\limits_{\substack{i \in V_L \\ i \neq k}}\sum\limits_{\substack{j \in N \\ \langle i, j, k \rangle \in \mathbb{P}}}y_{ijk} \leq 1 \quad \forall k \in V_R} \label{constraint:11}
\end{align}


\begin{align}
& \mathlarger{u_i - u_j \geq 1 - (n + 2)p_{ij} - M(2 - \sum\limits_{\substack{h \in V_L \\ h \neq i}}x_{hi} - \sum\limits_{\substack{k \in N \\ k \neq j}}x_{kj})} \label{constraint:12} \\
& \quad \forall i \in N, j \in \{V_R: j \neq i\} \nonumber \\
& \mathlarger{u_i - u_j \leq -1 + (n + 2)(1 - p_{ij}) + M(2 - \sum\limits_{\substack{h \in V_L \\ h \neq i}}x_{hi} - \sum\limits_{\substack{k \in N \\ k \neq j}}x_{kj})} \label{constraint:13} \\
& \quad \forall i \in N, j \in \{V_R: j \neq i\} \nonumber \\
& \mathlarger{u_0 - u_j \geq 1 - (n + 2)p_{0j} - M(1 - \sum\limits_{\substack{k \in V_L \\ k \neq j}}x_{kj}) \quad \forall j \in V_R} \label{constraint:14} \\
& \mathlarger{u_0 - u_j \leq -1 + (n + 2)(1 - p_{0j}) + M(1 - \sum\limits_{\substack{k \in V_L \\ k \neq j}}x_{kj}) \quad \forall j \in V_R} \label{constraint:15} \\
& \mathlarger{u_l \geq u_k - M \Bigg(3 - \sum\limits_{\substack{j \in N \\ j \neq l \\ \langle i, j, k \rangle \in \mathbb{P}}}y_{ijk} - \sum\limits_{\substack{m \in N \\ m \neq i \\ m \neq k \\ m \neq l}}\sum\limits_{\substack{n \in V_R \\ n \neq i \\ n \neq k \\ \langle l, m, n \rangle \in \mathbb{P}}}y_{lmn} - p_{il}\Bigg)} \label{constraint:16} \\
& \quad \forall i \in V_L, k \in \{V_R: k \neq i\}, l \in \{N: l \neq i, l \neq k\}. \nonumber
\end{align}

\begin{align}
& \mathlarger{t_k \geq r_i + \tau_{ik} - M(1 - x_{ik}) \quad \forall i \in V_L, k \in V_R, i \neq j} \label{constraint:17} \\
& \mathlarger{t_k \leq r_i + \tau_{ik} + M(1 - x_{ik}) \quad \forall i \in V_L, k \in V_R, i \neq j} \\
& \mathlarger{t'_j \geq r_i + \tau'_{ij} - M(1 - \sum\limits_{\substack{k \in V_R \\ \langle i, j, k \rangle \in \mathbb{P}}}y_{ijk}) \quad \forall	j \in V_D, i \in V_L, j \neq i} \\
& \mathlarger{t'_j \leq r_i + \tau'_{ij} + M(1 - \sum\limits_{\substack{k \in V_R \\ \langle i, j, k \rangle \in \mathbb{P}}}y_{ijk}) \quad \forall	j \in V_D, i \in V_L, j \neq i} \\
& \mathlarger{t'_k \geq r'_j + \tau'_{jk} - M(1 - \sum\limits_{\substack{i \in V_L \\ \langle i, j, k \rangle \in \mathbb{P}}}y_{ijk}) \quad \forall	j \in V_D, k \in V_R, j \neq k} \\
& \mathlarger{t'_k \leq r'_j + \tau'_{jk} + M(1 - \sum\limits_{\substack{i \in V_L \\ \langle i, j, k \rangle \in \mathbb{P}}}y_{ijk}) \quad \forall	j \in V_D, k \in V_R, j \neq k} \\
& \mathlarger{t'_j \geq r'_j - M (1 - \sum\limits_{\substack{i \in V_L \\ i \neq j}}\sum\limits_{\substack{k \in V_R \\ \langle i, j, k \rangle \in \mathbb{P}}}y_{ijk}) \quad \forall j \in N} \\
& \mathlarger{t'_j \leq r'_j + M (1 - \sum\limits_{\substack{i \in V_L \\ i \neq j}}\sum\limits_{\substack{k \in V_R \\ \langle i, j, k \rangle \in \mathbb{P}}}y_{ijk}) \quad \forall j \in N} \\
& \mathlarger{r_k \geq t_k + s_L (\sum\limits_{\substack{l \in N \\ l \neq k}}\sum\limits_{\substack{m \in V_R \\ m \neq l \\ m \neq k \\ \langle k, l, m \rangle \in \mathbb{P}}}y_{klm}) + s_R (\sum\limits_{\substack{i \in V_L \\ i \neq k}}\sum\limits_{\substack{j \in N \\ \langle i, j, k \rangle \in \mathbb{P}}}y_{ijk})} \\ \nonumber
& \mathlarger{\quad\quad\quad - M(1 - \sum\limits_{\substack{i \in V_L \\ i \neq k}}\sum\limits_{\substack{j \in N \\ \langle i, j, k \rangle \in \mathbb{P}}}y_{ijk}) \quad \forall k \in V_R} \\
& \mathlarger{r'_k \geq t'_k + s_L (\sum\limits_{\substack{l \in N \\ l \neq k}}\sum\limits_{\substack{m \in V_R \\ m \neq l \\ m \neq k \\ \langle k, l, m \rangle \in \mathbb{P}}}y_{klm}) + s_R (\sum\limits_{\substack{i \in V_L \\ i \neq k}}\sum\limits_{\substack{j \in N \\ \langle i, j, k \rangle \in \mathbb{P}}}y_{ijk})} \\ \nonumber
& \mathlarger{\quad\quad\quad - M(1 - \sum\limits_{\substack{i \in V_L \\ i \neq k}}\sum\limits_{\substack{j \in N \\ \langle i, j, k \rangle \in \mathbb{P}}}y_{ijk}) \quad \forall k \in V_R} \\
& \mathlarger{r'_k - (r'_j - \tau'_{ij}) - s_L (\sum\limits_{\substack{l \in N \\ l 
\neq i \\ l \neq j \\ l \neq k}}\sum\limits_{\substack{m \in V_R \\ m \neq k \\ m \neq i \\ m \neq l \\ \langle k, l, m \rangle \in \mathbb{P}}}y_{klm}) \leq \epsilon + M(1 - y_{ijk})} \\ \nonumber
& \mathlarger{\quad \forall k \in V_R, j \in C, j \neq k, i \in V_R, \langle i, j, k \rangle \in \mathbb{P}} \\
& \mathlarger{w_k \geq 0 \quad \forall k \in V_R} \\
& \mathlarger{w'_k \geq 0 \quad \forall k \in V_R} \\
& \mathlarger{w_k \geq t'_k - t_k \quad \forall k \in V_R} \\
& \mathlarger{w'_k \geq t_k - t'_k \quad \forall k \in V_R}  \\
& \mathlarger{w_0 = 0} \\
& \mathlarger{w'_0 = 0} \\
& \mathlarger{r_i = r'_i \quad \forall i \in V} \\
& \mathlarger{t_0 = 0} \\
& \mathlarger{t'_0 = 0} \\
& \mathlarger{r_0 = 0} \\
& \mathlarger{r'_0 = 0} \label{constraint:36}
\end{align}

\begin{align}
& x_{ij} \in \{0, 1\} \quad \forall i \in V_L, j \in V_R, j \neq i \\
& y_{ijk} \in \{0, 1\} \quad \forall i \in V_L, j \in N, k \in V_R, i \neq j, j \neq k, i \neq k, \langle i, j, k \rangle \in \mathbb{P} \\
& p_{ij} \in \{0, 1\} \quad \forall i, j \in N, i \neq j \\
& p_{0j} = 1 \quad \forall j \in N \\
& 0 \leq u_i \leq n + 1 \quad \forall i \in V \\
& t_i \geq 0 \quad \forall	i \in V \\
& t'_i \geq 0 \quad \forall	i \in V \\
& r_i \geq 0 \quad \forall	i \in V \\
& r'_i \geq 0 \quad \forall	i \in V 
\end{align}

The objective is to minimize the operational costs. We now explain the constraints. The letter in parenthesis at the end of each bullet item, if any, denotes the association between a MILP constraint and a constraint described in the model:
\begin{itemize}
\item Constraint \ref{constraint:2} guarantees that each node is visited once by either a truck or a drone. \textbf{(A)}
\item Constraints \ref{constraint:3} and \ref{constraint:4} state that the truck must start from and return to the depot. \textbf{(Modelling $TD$)}
\item Constraint \ref{constraint:5} is a subtour elimination constraint. \textbf{(Modelling $TD$)}
\item Constraint \ref{constraint:6} indicates that if the truck visits $j$ then it must depart from $j$. \textbf{(Modelling $TD$)}
\item Constraint \ref{constraint:7} associates a drone delivery with the truck route. In detail, if we have a drone delivery $\langle i, j, k \rangle$, then there must be a truck route between $i$ and $k$. \textbf{(C)}
\item Constraint \ref{constraint:8} indicates that if the drone is launched from the depot, then the truck must visit $k$ to collect it. \textbf{(C)}
\item Constraint \ref{constraint:9} ensures that if there is a drone delivery for $\langle i, j, k \rangle$, then the truck must visit $i$ before $k$. \textbf{(C)}
\item Constraints \ref{constraint:10} and \ref{constraint:11} state that each node in $V_L$ or $V_R$ can either launch the drone or retrieve it at most once, respectively. \textbf{(B)}
\item Constraints \ref{constraint:12}, \ref{constraint:13}, \ref{constraint:14} and \ref{constraint:15} ensure that if $i$ is visited before $j$ in the truck route, then its ordering constraint must be maintained. \textbf{(D)}
\item Constraint \ref{constraint:16}, if we have two drone deliveries $\langle i, j, k \rangle$ and $\langle l, m, n \rangle$ and $i$ is visited before $l$, then $l$ must be visited after $k$. This constraint avoids the problem of launching a drone between $i$ and $k$. \textbf{(D)}
\item Finally, constraints \ref{constraint:17} to \ref{constraint:36} ensure that waiting time and endurance is correctly handled. \textbf{(E)}
\end{itemize}

\section{A Greedy Randomized Adaptive Search Procedure (GRASP) for TSP-D}
\label{section:grasp}

This section presents a Greedy Randomized Adaptive Search Procedure (GRASP) \cite{gendreau2010handbook} to solve the min-cost TSP-D. We also adapt our split procedure to solve the min-time TSP-D. In the construction step, we propose a split algorithm that builds a min-cost TSP-D solution from a TSP solution. In the local search step, new operators adapted from the traditional ones are introduced for the min-cost TSP-D. The general outline of our GRASP is shown in Algorithm \ref{grasp-algorithm}. More specifically, in each iteration, it first generates a TSP tour using a TSP construction heuristic (line 7). In this paper, we use three heuristics to generate giant tours as follows:
\begin{itemize}
	\item \textit{$k$-nearest neighbour:} This heuristic is inspired from the well-known nearest neighbour algorithm for solving the TSP. It starts from the depot, repeatedly visits the node $v$ which is randomly chosen among $k$ closest unvisited nodes.
	\item \textit{$k$-cheapest insertion:} The approach is to start with a subtour, i.e., a small tour with a subset of nodes, and then extend this tour by repeatedly inserting the remaining nodes until no more node can be added. The unvisited node $v$ to be inserted and its insertion location between two consecutive nodes ($i$, $j$) of the tour are selected so that this combination gives the least Insertion Costs (IC). This cost is calculated by:
	\begin{equation}
		IC = d_{iv} + d_{vj} - d_{ij}
	\end{equation}
	To create the randomness for the heuristic, at each insertion step we randomly choose a pair of an unvisited node and its insertion location among $k$ pairs which provides the best insertion costs. The starting subtour includes only the depot.
	\item \textit{random insertion:} This heuristic works similarly to the $k$-nearest neighbour but it iteratively chooses a random node $v$ among all unvisited nodes.
\end{itemize}

\begin{algorithm}[H]
\KwResult{$bestSolution$}
$bestSolution$ = $null$ \;
$bestObjectiveValue$ = $\top$ \;
$randomGenerator$ = initialize TSP tour random generator \;
$iteration = 0$ \;
\While {$iteration < n_{TSP}$} {
$iteration = iteration + 1$ \;
tour = generate a random TSP tour using $randomGenerator$ \;
(P,V,T) = Split\_Algorithm\_Step1(tour) \;
tspdSolution = Split\_Algorithm\_Step2(P, V, T) \;
tspdSolution = Local\_Search(tspdSolution) \;
\If {f(tspdSolution) < bestObjectiveValue} {
$bestSolution$ = tspdSolution \;
$bestObjectiveValue$ = f(tspdSolution) \;
}
}
return $bestSolution$ \;
\caption{Greedy Randomized Adaptive Search Procedure (GRASP) for min-cost TSP-D} \label{grasp-algorithm} 
\end{algorithm}

In the next step, we construct a min-cost TSP-D solution using the split algorithm (line 8 and 9) and then improve it by local search (line 10). The best solution found is also recorded during the processing of the tours (lines 11 to 13). The algorithm stops after $n_{TSP}$ iterations. The detailed implementation of the split algorithm is described in Algorithms \ref{split-algorithm} and \ref{split-algorithm-get-solution}.

\subsection{A Split Algorithm for min-cost TSP-D}

Given a TSP tour, the split procedure algorithm selects nodes to be visited by the drone to obtain a solution for the min-cost TSP-D, assuming that the relative order of the nodes is fixed. Other split procedures are now used widely in state-of-the-art metaheuristics such as \cite{prins2009tour}, \cite{Vidal2013}, \cite{Vidal2014}, \cite{cattaruzza2014memetic} to solve many variants of VRPs. We start from a given TSP tour $s = (s_0, s_1, \ldots s_{n + 1})$ and must convert this tour into a feasible min-cost TSP-D solution. This is accomplished by removing nodes from the truck tour and substituting drone deliveries for those nodes. There are two main steps in the split algorithm: auxiliary graph construction and solution extraction. The pseudo code for these is listed in Algorithms \ref{split-algorithm} and \ref{split-algorithm-get-solution}, respectively. The most important step of the split algorithm is the construction of the auxiliary graph, in which each subsequence of nodes ($s_i, \ldots s_k$) can be turned into a drone delivery such that $s_i$ is the launch node, $s_k$ is the rendezvous node and $s_j$, where $pos(s_i, s)< pos(s_j, s) < pos(s_k, s)$, is the drone node. We now describe the split algorithm in detail. \\

\scalebox{0.84} {
\begin{algorithm}[H]
\KwData{TSP tour $s$}
\KwResult{P stores the shortest path from the auxiliary graph, V is the cost of that shortest path, and T is a list of the possible drone deliveries and costs}

$arcs$ = $\emptyset$ \; 
T = $\emptyset$ \;
\textit{/* Auxiliary graph construction - Arcs */} \\
\ForEach{ i in s $\setminus$ last(s)} {
$k = pos(i, s) + 1$ \;
$arcs$ = $arcs$ $\cup$ ($i, k, cost(i, k, s)$)
}
\ForEach { i in s $\setminus$ \{ last(s), s[pos(last(s), s) - 1] \} } {
\ForEach { k in s : pos(k, s) $\geq$ pos(i, s) + 2} {
minValue = $\infty$ \;
minIndex = $\infty$ \;
\ForEach { j in s : pos(i, s) < pos(j, s) < pos(k, s) } {
\If {$\langle i, j, k \rangle \in \mathbb{P}$} {
cost = $cost(sub(i, k, s)) + C_1\Big( d_{prev_s(j) next_s(j)} - d_{prev_s(j),j} - d_{j,next_s(j)} \Big) + cost(i, j, k) + cost^T_W(i,j,k) + cost^D_W(i,j,k)$ \;
\If { cost < minValue } {
minValue = cost \; 
minIndex = pos(j, s) \; 
}
}
}
$arcs$ = $arcs$ $\cup$ $\{(i, k, minValue)\}$ \;
\If {minIndex $\neq$ $\top$} {
T = T $\cup$ $\{(i, s[minIndex], k, minValue)\}$ \;
}
} 
}
\textit{/* Finding the shortest path */} \\
V[0] = 0 \;
P[0] = 0 \;
\ForEach { k in s $\setminus$ \{ 0 \}} {
\ForEach {($i, k, cost$) $\in$ $arcs$} {
\If {V[k] > V[i] + $cost$} {
V[k] = V[i] + $cost$ \;
P[k] = i \;
}
}
}
return (P, V, T) \;
\caption{Split\_Algorithm\_Step1(s): Building the auxiliary graph and finding shortest path} \label{split-algorithm} 
\end{algorithm}
}

\paragraph{Building the auxiliary graph and finding shortest path}

In Algorithm \ref{split-algorithm}, we construct an auxiliary weighted graph $H = (V', A')$ based on the TSP tour $s$ of the graph $G = (V, A)$. We have $V' = V$ and an arc ($i, j$) $\in A'$ that represents a subroute from $i$ to $j$, where $pos(i, s) < pos(j, s)$. 

If $i$ and $j$ are adjacent nodes in $s$, then the cost $c_{ij}$ of arc ($i, j$) $\in A'$ is calculated directly as follows:
\begin{equation}
c_{ij} = C_1 d_{ij}.
\end{equation}

However, when $i$ and $k$ are not adjacent and a node $j$ exists between $i$ and $k$ such that $\langle i, j, k \rangle \in \mathbb{P}$, then 
\begin{small}
\begin{equation}
\begin{split}
c_{ik} = \min_{\langle i, j, k \rangle \in \mathbb{P}} cost(sub(i, k, s)) + C_1\Big( d_{prev_s (j), next_s (j)} - d_{prev_s (j) ,j} - d_{j, next_s (j)} \Big) \\ + cost(i, j, k) + cost^T_W(i,j,k) + cost^D_W(i,j,k).
\end{split}
\end{equation}
\end{small}

If $i$ and $k$ are not adjacent and no node $j$ exists between $i$ and $k$ such that $\langle i, j, k \rangle$ could be a \textbf{drone delivery}, then
\begin{equation}
c_{ik} = +\infty.
\end{equation}

The arc's cost calculation is shown in lines 1 to 19 in Algorithm \ref{split-algorithm}. Moreover, in lines 18 and 19, we store the list of possible \textbf{drone deliveries} T. This list will be used in the extraction step. 

The auxiliary graph is used to compute the cost $v_k$ of the shortest path from the depot to node $k$. Because the graph $H$ is a directed acyclic graph, these values can be computed easily using a dynamic programming approach. Moreover, an arc ($i, k$) in the shortest path that does not belong to the initial TSP tour means that a drone delivery can be made where $i$ is the launch node, $k$ is the rendezvous node, and the delivery node is a node between $i$ and $k$ in the TSP tour. This computation ensures that no interference occurs between the chosen drone deliveries. We therefore obtain the best solution from the TSP tour while respecting the relative order of the nodes.

In detail, given $v_0 = 0$, the value $v_k$ of each node $k \in V' \setminus \{0\}$ is then calculated by 
\begin{equation}
v_k = min \{ v_{i} + c_{ik} : (i, k) \in A' \} \quad \forall k = 1,2,\ldots, n + 1.
\end{equation}
We also store the shortest path from $0$ to $n + 1$ in $P(j)$, where $j = 1 \ldots n + 1$; $j$ is the node, and the value $P(j)$ is the previous node of $j$. These steps are described in lines 21 to 27 in Algorithm \ref{split-algorithm}. An auxiliary graph for Figure \ref{figure:tsp-tspd} is shown in Figure \ref{figure:split-auxiliary-graph}

\begin{small}
\begin{figure}[H]
\centering
\scalebox{0.8}{
\begin{tikzpicture}[
> = stealth, 
shorten > = 1pt, 
auto,
node distance = 2cm, 
semithick 
]
\tikzstyle{every state}=[
draw = black,
thick,
fill = white,
minimum size = 4mm
]

\node[state] (0) {$0$};
\node[state] (6) [right of=0] {$6$};
\node[state] (3) [right of=6] {$3$};
\node[state] (5) [right of=3] {$5$};
\node[state] (2) [right of=5] {$2$};
\node[state] (1) [right of=2] {$1$};
\node[state] (4) [right of=1] {$4$};
\node[state] (7) [right of=4] {$7$};
\path[->] (0) edge node {225} (6);
\path[->] (0) edge [bend right=70] node [pos=0.5] {413} (3);
\path[->] (0) edge [bend right=70] node [pos=0.5] {$\infty$} (5);
\path[->] (0) edge [bend right=70] node [pos=0.5] {$\infty$} (2);
\path[->] (0) edge [bend right=70] node [pos=0.5] {1264} (1);
\path[->] (0) edge [bend right=70] node [pos=0.5] {1258} (4);
\path[->] (0) edge [bend right=70] node [pos=0.5] {1464} (7);
\path[->] (6) edge node {175} (3);
\path[->] (6) edge [bend left=70] node [pos=0.6] {390} (5);
\path[->] (6) edge [bend left=70] node [pos=0.6] {$\infty$} (2);
\path[->] (6) edge [bend left=70] node [pos=0.6] {938} (1);
\path[->] (6) edge [bend left=70] node [pos=0.6] {989} (4);
\path[->] (6) edge [bend left=70] node [pos=0.6] {1192} (7);
\path[->] (3) edge node {300} (5);
\path[->] (3) edge [bend right=60] node [pos=0.5] {569} (2);
\path[->] (3) edge [bend right=60] node [pos=0.6] {$\infty$} (1);
\path[->] (3) edge [bend right=60] node [pos=0.5] {859} (4);
\path[->] (3) edge [bend right=60] node [pos=0.5] {1065} (7);
\path[->] (5) edge node {250} (2);
\path[->] (5) edge [bend left=60] node [pos=0.5] {368} (1);
\path[->] (5) edge [bend left=60] node [pos=0.6] {419} (4);
\path[->] (5) edge [bend left=60] node [pos=0.6] {767} (7);
\path[->] (2) edge node {300} (1);
\path[->] (2) edge [bend right=45] node [pos=0.6] {310} (4);
\path[->] (2) edge [bend right=45] node [pos=0.7] {516} (7);
\path[->] (1) edge node {50} (4);
\path[->] (1) edge [bend left=45] node [pos=0.5] {257} (7);
\path[->] (4) edge node {200} (7);
\end{tikzpicture}
}
\caption{Auxiliary graph for TSP tour in Figure \ref{figure:tsp-tspd}} \label{figure:split-auxiliary-graph}
\end{figure}
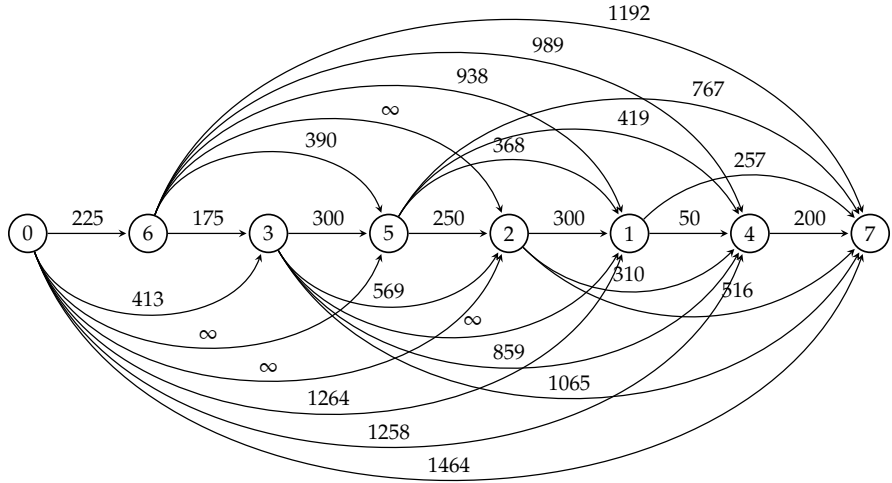
\end{small}

The cost computation for each arc in the graph $H$ can be done in $\mathcal{O}(n^2)$. Since $H$ is acyclic by construction, to search for a shortest path, a breadth-first search (BFS) algorithm for directed acyclic graphs can be used, with an $\mathcal{O}(|A|)$ complexity where $|A|$ is the number of arcs in the graph. Because the number of arcs in $H$ is proportional to $n^2$. Thus, the search for a shortest path in graph $H$ can be done in $\mathcal{O}(n^2)$. Several Split procedures in the literature work in a similar manner (see \cite{prins2004simple}, \cite{prins2009grasp} for example).  Therefore, we get the complexity of the Algorithm \ref{split-algorithm} in $\mathcal{O}(n^4)$.

\paragraph{Extracting min-cost TSP-D solution}

Given $P(j), j = 1 \ldots n + 1$ defined as above and a list of possible \textbf{drone deliveries} T, we now extract the min-cost TSP-D solution in Algorithm \ref{split-algorithm-get-solution}. In the first step, given $P$, we construct a sequence of nodes $S_a = 0, n_1, \ldots, n + 1$ representing the path from 0 to n + 1 in the auxiliary graph (lines 2 to 9). Each two consecutive nodes in $S_a$ are a subroute of the complete solution. However, they might include a \textbf{drone delivery}; consequently, we need to determine which node might be the drone node in the subroute, which is computed in $T$.

The second step is to construct a min-cost TSP-D solution. To do that, we first initialize two empty sets: a set of drone deliveries $S_d$ and a set representing the truck's tour sequence $S_t$ (lines 11 and 12). We now build these sets one at a time.

For drone delivery extractions, we consider each pair of adjacent positions $i$ and $i + 1$ in $P_{new}$ and determine the number of in-between nodes. If there is at least one $j$ node between the $i$ and $i + 1$ positions in the TSP tour, we will choose the drone delivery in T with the minimum value, taking its drone node $j$ as the result (lines 14 to 17).

To extract the truck's tour (line 19 to 26), we start from the depot $0$ in $S_a$. Each pair $i, i + 1 \in S_a$ is considered as a subroute in the min-cost TSP-D solution by taking the nodes from $i$ to $i + 1$ in the TSP solution. However, in cases where $i$ and $i + 1$ are launch and rendezvous nodes of a \textbf{drone delivery}, respectively, $\langle i, j, i + 1 \rangle$, $j$ must not be considered in the truck's tour. \\

\scalebox{0.89} {
\begin{algorithm}[H]
\KwData{P stores the path in the auxiliary graph, V is the cost of the path in P, T is the list of drone deliveries + costs, and $tspTour$ is the truck-only TSP tour}
\KwResult{$tspdSolution$}

\textit{/* Construct the sequence of nodes representing the path stored in P */} \\
j = n + 1 \;
i = $\infty$ \;
$S_a$ = $\langle$ j $\rangle$ \;
\While{ i $\neq$ 0 } {
i = P[j] \;
$S_a$ = $S_a$::$\langle$i$\rangle$ \;
j = i \;
}
$S_a$ = $S_a.reverse()$ \;
\textit{/* Create a min-cost TSP-D solution from $S_a$ */} \\
$S_d$ = $\langle\rangle$ \;
$S_t$ = $\langle\rangle$ \;
\textit{/* Drone deliveries */} \\
\For {i = 0; i < $S_a.size$ - 1; i++} {
\If {between $S_a[i]$ and $S_a[i + 1]$ in $tspTour$, there is at least one node} {

$n_{drone}$ = obtain the associated drone node in tuples T \;
$S_d$ = $S_d$ $\cup$ $\langle S_a[i], n_{drone}, S_a[i + 1] \rangle$ \;
}
}
\textit{/* Truck tour */} \\
currentNode = 0 \;
\While {currentNode $\neq$ n + 1} {
\If {currentNode is a launch node of a tuple $t$ in $S_d$} {
$S_t$ = $S_t$ :: $\langle$all the nodes from the currentNode to the return node of $t$ in $tspTour$ except the drone node$\rangle$ \;
currentNode = the return node of $t$ \;
}
\Else {
$S_t$ = $S_t$ :: $\langle$currentNode$\rangle$ \;
currentNode = $tspTour[indexOf(currentPosition) + 1]$ \;
}
}
$tspdSolution = (S_t, S_d)$ \;
return $tspdSolution$ \;
\caption{Split\_Algorithm\_Step2(P,V,T): Extract\_TSPD\_Solution} \label{split-algorithm-get-solution} 
\end{algorithm}
}

\paragraph{Split procedure adaptation for min-time TSP-D} To deal with the min-time problem, we change the way the arc's costs are computed in the auxiliary graph as follows:
If $i$ and $j$ are adjacent nodes in $s$, then the cost $c_{ij}$ is calculated by:
\begin{equation}
c_{ij} = \tau_{ij}
\end{equation}
When $i$ and $k$ are not adjacent and a node $j$ exists between $i$ and $k$ such that $\langle i, j, k \rangle \in \mathbb{P}$, then 
\begin{equation}
\begin{multlined}
c_{ik} = \min_{\langle i, j, k \rangle \in \mathbb{P}} ( max(time_T(i \rightarrow k), time_D(i, j) + time_D(j, k)) + s_R + s_L).
\end{multlined}
\end{equation}

where $time_T(i \rightarrow k)$ is the travel time of truck from launch point $i$ to rendezvous point $j$ and $time_D(i, j)$ is the travel time of drone from $i$ to $j$. Eventually, this modification results in a change of Algorithm \ref{split-algorithm}, specifically, line 13 to 15 as follows:

\scalebox{0.84} {
\begin{algorithm}[H]
...\

$timeDrone = time_D(i, j) + time_D(j, k)$\;
$timeTruck = time_T(i \rightarrow k)$\;
\If { $max(timeTruck, timeDrone) + s_R + s_L$ < minValue } {
minValue = $max(timeTruck, timeDrone) + s_R + s_L$ \; 

...\
}
\caption{Split\_Algorithm\_Step1(s): Min-time adaptation} \label{split-algorithm-min-time} 
\end{algorithm}
}

\subsection{Local search operators}

Two of our local search operators are inspired from the traditional move operators Two-exchange and Relocation \cite{kindervater1997vehicle}. In addition, given the characteristics of the problem, we also develop two new move operators, namely, "drone relocation", which is a modified version of the classical relocation operator, and "drone removal", which relates to the removal of a drone node. In detail, from a min-cost TSP-D solution $(TD, DD)$, we denote the following:
\begin{itemize}
\item[-] $N_T(TD, DD) = \{ e: e \in TD, \langle e, \cdot, \cdot \rangle \notin DD, \langle \cdot, \cdot, e \rangle \notin DD \}$ is the set of \textbf{truck-only nodes} in the solution $(TD, DD)$ that are not associated with any drone delivery
\item[-] $N_D(TD, DD) = \{ e: \langle \cdot, e, \cdot \rangle \in DD \}$ is the set of \textbf{drone nodes} in the solution $(TD, DD)$
\end{itemize}
We now describe each operator.

\begin{figure}[h!] 
\centering
\includegraphics[scale=0.6]{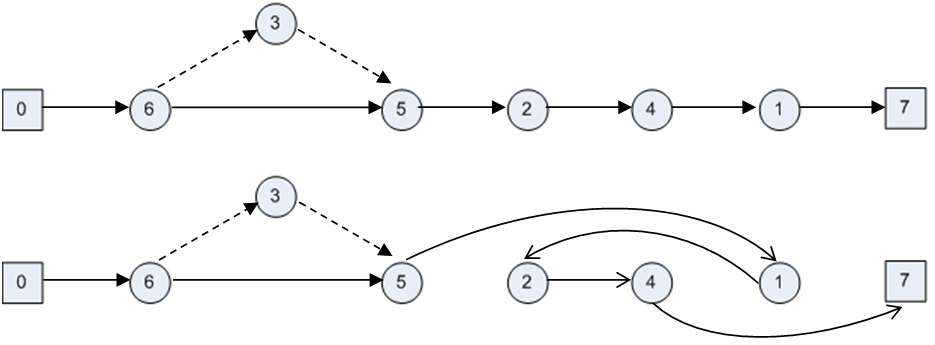} 
\caption{A truck relocation move operator : $relocate_T((TD, DD), 1, 2)$}\label{figure:ls-truck-relocation}
\end{figure}

\textit{Relocation}: This is the traditional relocation operator with two differences: (1) We consider only truck-only nodes; (2) we only relocate into a new position in the truck's tour. An example is shown in Figure \ref{figure:ls-truck-relocation}. In detail, we denote
\begin{equation}
relocate_{T}((TD, DD), a, b), 
a \in N_T((TD, DD)),
b \in TD, b \neq a, b \neq 0
\end{equation}
as the operator that---in effect---relocates node $a$ before node $b$ in the truck tour.

\begin{figure}[h!] 
\centering
\includegraphics[scale=0.6]{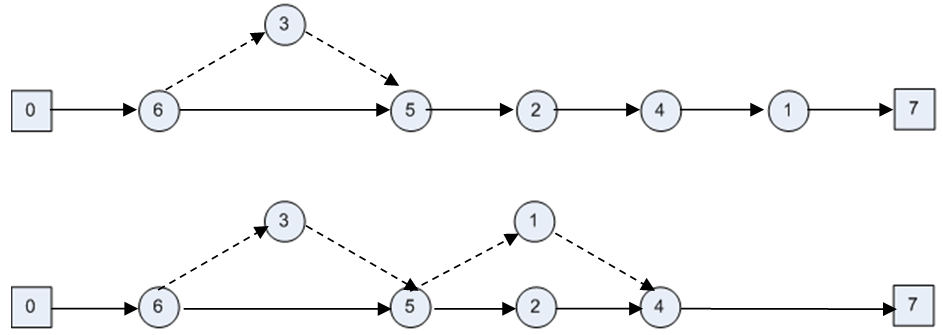} 
\caption{A drone relocation move operator : $relocate_D((TD, DD),1, 5, 4)$}\label{figure:ls-drone-insertion}
\end{figure}

\textit{Drone relocation}: The original idea of this operator is that it can change a truck node to a drone node or relocate an existing drone node so that it has different launch and rendezvous locations. The details are as follows: (1) We consider both truck-only and drone nodes; (2) each of these nodes is then relocated as a drone node in a different position in the truck's tour. This move operator results in a neighbourhood that might contain more \textbf{drone deliveries}; hence, it has more possibilities to reduce the cost. An example is shown in Figure \ref{figure:ls-drone-insertion}. More precisely, we denote 
\begin{equation}
relocate_{D}((TD, DD), a, i, k)
\end{equation}
\begin{equation}
a \in N_T((TD, DD)) \cup N_D((TD, DD)),
i, k \in TD \setminus \{a\}, i \neq k,
\nonumber
\end{equation}
\begin{equation}
pos(i, TD) < pos(k, TD),
\langle i, a, k \rangle \in \mathbb{P} \nonumber
\end{equation}
as the operator procedure, where $a$ is the node to be relocated and $i$ and $k$ are two nodes in $TD$. There are two possibilities for effects: (1) If $a$ is a truck-only node, this move creates a new drone delivery $\langle i, a, k \rangle$ in $DD$ and removes $a$ from $TD$; (2) if $a$ is a drone node, the move changes the drone delivery $\langle \cdot, a, \cdot \rangle \in DD$ to $\langle i, a, k \rangle$.

\begin{figure}[h!] 
\centering
\includegraphics[scale=0.6]{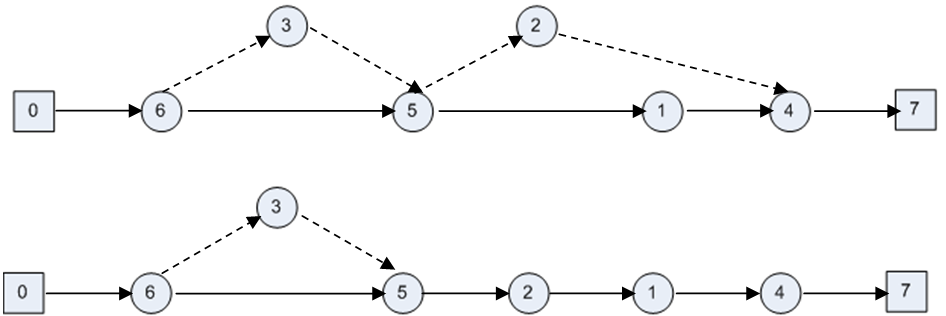} 
\caption{A drone removal move operator : $remove_D((TD, DD), 2)$}\label{figure:ls-drone-removal}
\end{figure}

\textit{Drone removal}: In this move operator, we choose a drone node $j \in N_D$ and replace the drone delivery by a truck delivery. An example is shown in Figure \ref{figure:ls-drone-removal}. In detail, we denote
\begin{equation}
remove_D((TD, DD), j, k),
j \notin TD, \langle \cdot, j, \cdot \rangle \in DD, k \in TD, k \neq \{0\}
\end{equation}
as the operator procedure, where $j$ is the drone node to be removed and $k$ is a node in $TD$ such that $j$ will be inserted before $k$. As a result, we have a new solution in which the number of nodes in $TD$ has been increased by one and $DD$'s cardinality has been decreased by one.

\begin{figure}[h!] 
\centering
\includegraphics[scale=0.7]{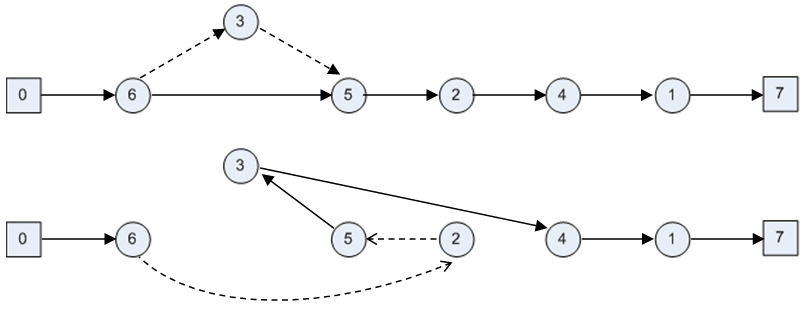} 
\caption{A two-exchange move operator in which a drone node is exchanged with a truck node : $two\_exchange((TD, DD), 3, 2)$}\label{figure:ls-two-exchange}
\end{figure}

\textit{Two-exchange}: We exchange the position of two nodes. There are three possibilities: (1) When the exchanged nodes are both drone nodes, we make the change in the drone delivery list; (2) when the exchanged nodes are both truck nodes, we first exchange their positions in the truck sequence and then apply changes in the drone delivery list; and (3) when the exchanged nodes are a truck node and a drone node, we remove the old tuple and create a new one with the exchanged node. Next, we update the truck sequence and apply the changes to the tuples if the truck node is associated with any drone delivery. An example is shown in Figure \ref{figure:ls-two-exchange}. In detail, we denote
\begin{equation}
two\_exchange((TD, DD), a, b), a, b \in V \setminus \{0, n + 1\}, a \neq b
\end{equation}
as the operator procedure, where $a$ and $b$ are the two nodes to be exchanged. We then swap their positions. The three swap possibilities are as follows: (1) a drone node with a node in $TD$; (2) two drone nodes; and (3) two nodes in $TD$.

To ensure the feasibility of resulting solutions, we only accept the moves which satisfy the constraints of the problem. And finally, our local search operators are easily adapted to deal with the min-time objective. They work on the travel time instead of the travel cost of each arc. Whenever there is a need to update a travel time of a drone delivery, we need to take the greater value between travel times of drone and truck instead of the summation. The mechanism of the rest of the local search then stays untouched.

\section{TSP-LS heuristic}
\label{section:TSP-LS}

The TSP-LS algorithm is adapted from the work of \cite{murray2015flying} to solve the min-cost TSP-D. The differences between min-time FSTSP and the adapted min-cost TSP-LS are at the calculation of cost savings (Algorithm \ref{TSP-LS-calc-savings}), the cost of relocating a truck node to another position (Algorithm \ref{TSP-LS-relocateAsTruck}) and the cost of inserting a node as a drone node between two nodes in the truck tour (Algorithm \ref{TSP-LS-relocateAsDrone}). These changes are not only about the unit of measurement (time vs. cost) but also the waiting cost of two vehicles. We now describe the algorithm in details.

The algorithm starts by calculating a TSP tour and then repeatedly relocates customers until no more improvement can be reached. The outline is shown in Algorithm \ref{TSP-LS-heuristic}. Lines 1--8 define the global variables, which are $Customers = [1, 2, \ldots, n]$, the sequence of truck nodes---$truckRoute$, and an indexed list $truckSubRoutes$ of smaller sequences that represent the subroutes in $truckRoute$. The distinct combination of elements in $truckSubRoutes$ must be equal to $truckRoute$. We define $i^*, j^*, k^*$, where $j^*$ is the best candidate for relocation and $i^*$ and $k^*$ denote the positions between which $j^*$ will be inserted. We also store $maxSavings$ which is the cost improvement value of this relocation. The two Boolean variables $isDroneNode$ and $Stop$ respectively determine whether a node in a subroute is a drone node and whether TSP-LS should terminate. These global variables are updated during the iterations, and the heuristic terminates when no more positive $maxSavings$ can be achieved ($maxSavings = 0$). \\

\begin{algorithm}[H] 
\KwData{truck-only sequence $truckRoute$}
\KwResult{TSP-D solution $sol$}
$Customers = N$ \;
$truckRoute$ = $solveTSP(N)$\;
$truckSubRoutes = \{truckRoute\}$\;
$sol$ = ($truckRoute$, $\emptyset$)\; 
$i^* = -1$\;
$j^* = -1$\;
$k^* = -1$\;
$maxSavings$ = 0\;
$isDroneNode = null$\;
$Stop = false$\;
\Repeat{Stop} {
\ForEach{$j \in Customers$} {
$savings = $ $calcSavings(j)$ \;
\ForEach{$subroute$ in $truckSubRoutes$} {
\If{drone(subroute, $sol$)} {
($isDroneNode, maxSavings, i^*, j^*, k^*$) = $relocateAsTruck(j, subroute, savings)$\;
}
\Else {
($isDroneNode, maxSavings, i^*, j^*, k^*$) = $relocateAsDrone(j, subroute, savings)$\;
}
}
}
\If {$maxSavings > 0$} {
($sol, truckRoute, truckSubRoutes, Customers$) = applyChanges($isDroneNode, i^*, j^*, k^*,$\
$sol, truckRoute, truckSubRoutes, Customers$)\;
$maxSavings = 0$ \;
}
\Else {
$Stop = true$\;
}
}
return $truckSubRoutes$\;
\caption{TSP-LS heuristic} \label{TSP-LS-heuristic} 
\end{algorithm}

For an additional notation used in Algorithm \ref{TSP-LS-heuristic}, line 15, given a solution $(TD, DD)$, we denote $drone(s, (TD, DD)) \in \{True, False\}$ as $True$ if the subsequence $s$ in $TD$ is associated with a drone:

\[ drone(s, (TD, DD)) = \left\{ \begin{array}{ll}
True & \mbox{if $\exists j \in V(s), j \neq first(s),$}\\
 & \mbox{$j \neq last(s) : \langle first(s), j, last(s) \rangle \in DD;$} \\
False & \mbox{if $\forall j \in V(s), j \neq first(s),$} \\
 & \mbox{$ j \neq last(s) : \langle first(s), j, last(s) \rangle \notin DD$}. 
 \end{array} \right. \] 

In detail, each iteration has two steps: (1) Consider each customer in $Customers$ to determine the best candidate for relocation along with its new position and the cost savings. (2) If the candidate relocation can improve the current solution, then relocate the customer by updating $truckRoute$ and $truckSubRoutes$ and remove it from $Customers$ so that it will not be considered in future iterations; otherwise (when the candidate relocation cannot improve the current solution), the relocation terminates. We now explain each step and its implementations in Algorithms \ref{TSP-LS-heuristic}, \ref{TSP-LS-calc-savings}, \ref{TSP-LS-relocateAsTruck}, \ref{TSP-LS-relocateAsDrone}, \ref{TSP-LS-applyChanges}.

Step 1 of the iteration is presented from lines 12 to 18 in Algorithm \ref{TSP-LS-heuristic}. It first considers each customer $j$ (line 12) and then calculates the cost savings by removing $j$ from its current position (line 13). The calculation is shown in Algorithm \ref{TSP-LS-calc-savings}. Next, line 14 considers each subroute in $truckSubRoute$ as a possible target for the relocation of $j$. When the current considered subroute is a \textbf{drone delivery} (line 15), we then try to relocate $j$ into this subroute as a truck node (line 16); otherwise, we try to relocate $j$ as a drone node to create a new \textbf{drone delivery} (line 18). The relocation analyses of $j$ as a truck node or a drone node are presented in Algorithms \ref{TSP-LS-relocateAsTruck} and \ref{TSP-LS-relocateAsDrone}, respectively. \\

\begin{algorithm}[H]
\KwData{$j$ : a customer currently assigned to the truck}
\KwResult{Solution}
$i = prev_{truckRoute}(j)$ \;
$k = next_{truckRoute}(j)$ \;
$savings = ( d_{i,j} + d_{j,k} - d_{i,k}) C_1$ \;
\If {j is associated with a drone delivery in subroute $s$} {
	$i = first(s)$\;
	$k = last(s)$\;
	$w = \alpha \times max(0, t_{i \rightarrow k} - \tau_{ij} - \tau_{jk} + \tau_{ik} - t'_{ijk})$ \;
	$w' = \beta \times max(0, t'_{ijk} - (t_{i \rightarrow k} - \tau_{ij} - \tau_{jk} + \tau_{ik}))$ \;
	$savings = savings + w + w'$
}
return $savings$\;
\caption{calcSavings(j)} \label{TSP-LS-calc-savings} 
\end{algorithm}

In Algorithm \ref{TSP-LS-relocateAsTruck}, we aim to find the best position in subroute $s$ to insert the current customer under consideration $j$ by checking each pair of adjacent nodes $i$ and $k$ in $s$ (line 3). After that, if the cost of inserting $j$ in this position is less than the current $savings$, then relocating $j$ here results in some savings (line 5). Furthermore, because this subroute has a \textbf{drone delivery}, we need to check whether inserting $j$ into it still lies within the drone's power limit so that the truck can still pick up the drone (line 6). Finally, if the cost saved is below the best known $maxSavings$, we apply the changes to this location by updating the values of $isDroneNode, i^*, j^*, k^*$ and $maxSavings$ (lines 7 to 10). \\

\begin{algorithm}[H] 
\KwData{\\ $j$ : current customer under consideration\\ $s$ : current subroute under consideration\\ $savings$ : savings that occur if $j$ is removed from its current position} 
\KwResult{Updated $i^*$, $j^*$, $k^*$, $isDroneNode$}
$a = first(s)$ \;
$b = last(s)$ \;
\ForEach{$(i, k) \in A(s)$} {
$\Delta = (d_{i,j} + d_{j,k} - d_{i,k}) C_1$ \;
\If {$\Delta < savings$} {
\If {the drone is still feasible to fly} {
\If {$savings - \Delta > maxSavings$} {
$isDroneNode = False$\;
$j^{*} = j; i^{*} = i; k^{*} = k$\;
$maxSavings = savings - \Delta$\;
}
}
}
}
return ($isDroneNode, maxSavings, i^*, j^*, k^*$)\;
\caption{relocateAsTruck(j, subroute, savings)---Calculates the cost of relocating the customer $j$ into a different position in the truck's route} \label{TSP-LS-relocateAsTruck} 
\end{algorithm}

In Algorithm \ref{TSP-LS-relocateAsDrone}, we consider the relocation of a customer $j$ in a subroute $s$ that does not have \textbf{drone delivery}. The objective is simple: try to make $j$ become the drone node of this subroute to reduce the cost. Hence, we consider each pair of $i$ and $k$ in $s$, where $i$ precedes $k$ (line 1, 2), and check whether $\langle i, j, k \rangle$ could be a viable \textbf{drone delivery} (line 3). We then calculate the cost of this change in lines 4--6. Next, we check whether the relocation is better than the best known $maxSavings$ in line 7. Finally, we update the relocation information in lines 8--10 as in Algorithm \ref{TSP-LS-relocateAsTruck} \\

\begin{algorithm}[H]
\KwData{\\ $j$ : current considered customer \\ $s$ : current considered subroute \\ $savings$ : current savings if $j$ is removed from its position}
\KwResult{Updated $i^*$, $j^*$, $k^*$, $isDroneNode$}
\For{$i = 0$ to $size(s) - 2$} {
\For{$k = i + 1$ to $size(s) - 1$} { 
\If {$\langle s[i], j, s[k] \rangle \in \mathbb{P}$} {
$w_k = $ waiting cost of truck at $k$ if $j$ is drone node \;
$w'_k = $ waiting cost of drone at $k$ if $j$ is drone node \;
$\Delta =(d'_{s[i],j} + d'_{j,s[k]}) C_2 + w_k + w'_k$ \;
\If {$savings - \Delta > maxSavings$} {
$isDroneNode = True$\;
$j^{*} = j; i^{*} = s[i]; k^{*} = s[k]$\;
$maxSavings = savings - \Delta$\;
}
} 
}
}
return ($isDroneNode, maxSavings, i^*, j^*, k^*$)\;
\caption{relocateAsDrone(j, subroute, savings) - Calculates the cost of relocating customer $j$ as a drone node} \label{TSP-LS-relocateAsDrone} 
\end{algorithm}

In step 2 of the iteration in Algorithm \ref{TSP-LS-applyChanges}, when any cost reduction exists ($maxSavings \neq 0$), we apply the changes based on the current values of $i^*, j^*, k^*, and isDroneNode$. If $isDroneNode = True$, we relocate $j^*$ between $i^*$ and $k^*$ as a drone node, forming a \textbf{drone delivery} (line 1 to 5). Otherwise, $j^*$ is inserted as a normal truck node (line 6 to 8). More specifically, these changes take place on the $truckRoute$ and $truckSubRoutes$.

Returning to Algorithm \ref{TSP-LS-heuristic}, after the changes have been applied in line 18, we reset the value of $maxSavings$ to 0 to prepare for the next iteration. Moreover, the algorithm terminates when $maxSavings = 0$ (line 21). \\

\begin{algorithm}[H] 
\KwData{$isDroneNode$, $i^{*}$, $j^{*}$, $k^{*}$, $sol$, $truckRoute$, $truckSubRoutes$, $Customers$}
\KwResult{Updated $truckRoute$, $truckSubRoutes$, $t$}
\If {$isDroneNode == True$} {
The Drone is now assigned to $i^{*} \rightarrow j^{*} \rightarrow k^{*}$\;
Remove $j^{*}$ from $truckRoute$ and $truckSubRoutes$\;
Append a new truck subroute that starts at $i^{*}$ and ends at $k^{*}$\;
Remove $i^{*}$, $j^{*}$, $k^{*}$ from $Customers$\;
}
\Else {
Remove $j^{*}$ from its current truck subroute\;
Insert $j^{*}$ between $i^{*}$ and $k^{*}$ in the new truck subroute \;
}
Update $sol$ using $truckRoute$ and $truckSubRoutes$ \; 
return ($sol, truckRoute, truckSubRoutes, Customers$)\;
\caption{applyChanges function} \label{TSP-LS-applyChanges} 
\end{algorithm}

\section{Experiment setup}

\label{section:experiments}

For the experiments, we generate customer locations randomly on a plane. We consider graphs with 10, 50 and 100 customers. These customers are created in squares with three different areas: 100~$km^2$, 500~$km^2$ and 1000~$km^2$. An instance of the TSP-D is characterised by: customer locations, total area of the plane, drone endurance, depot location as well as speed, distance types, travelling cost and time of each vehicle, drone launch time and retrieve time. In total, 65 instances are generated; their characteristics are partially shown in Table \ref{table:instances}:

\begin{table}[H]
\begin{center}
\scalebox{0.8}{
\begin{tabular}{ | l | c | c | c | c | c |}
\hline
Instances & \# of Customers & Area ($km^2$) & Density & Distance (km) & |$\mathbb{P}$| \\ \hline
A1 to A5 & 10 & 100 & 1 & 7.43 & 595 \\
B1 to B10 & 50 & 100 & 0.5 & 7.13 & 73053 \\
C1 to C10 & 50 & 500 & 0.1 & 15.45 & 10005 \\
D1 to D10 & 50 & 1000 & 0.05 & 22.19 & 2932 \\ \hline
E1 to E10 & 100 & 100 & 1 & 7.14 & 590144 \\
F1 to F10 & 100 & 500 & 0.2 & 15.21 & 81263 \\
G1 to G10 & 100 & 1000 & 0.1 & 21.59 & 24666 \\
\hline
\end{tabular}
}
\caption{Instances of min-cost TSP-D}
\label{table:instances}
\end{center}
\end{table}

The numbers in this table represent the average values over each class of instances. Three first columns "Instances", "\# of Customers", and "Area" are self-explained. Column "Density" represents the number of customers generated in an area unit while column "Distance" indicates the average Euclidean distance among customers. And finally, column "|$\mathbb{P}$|" implies the number of possible drone deliveries.

For all instances, the speeds of drone and truck are both set to 40~km/h. Moreover, $d_{ij}$ is calculated using Manhattan distance, while $d'_{ij}$ is in Euclidean distance. The objective here is to partially simulate the fact that the truck has to travel through a road network (which is longer) and the drone can fly directly from an origin to a destination. The drone's endurance $\epsilon$ is set to 20 minutes of flight time. The truck's cost $C_1$ is by default set to 25 times the drone's cost $C_2$. Depot location is at the bottom left of the square. To simulate the real situation where not all packages can be delivered by drone, in all instances, only 80~\% of customers can be served by drone. Waiting penalty coefficients $\alpha$ and $\beta$ are set to 10. And finally, the launch time $s_L$ and retrieve time $s_R$ are all set to 1 minute, as in \cite{murray2015flying}.

For the results, we denote $\gamma, T$, and $\rho$ as the objective value, running time in seconds and performance ratio, respectively, defined as follows:
\begin{equation}
\rho = \frac{value}{referenceValue} \times 100,
\end{equation}
where $value$ is the objective value obtained by the considering algorithm and $referenceValue$ is the objective value obtained by a reference algorithm. We will specify these algorithms for each experiment. Because we are dealing with a minimization problem, a ratio $\rho$ less than 100~\% means that the considered algorithm provides a better solution than the reference algorithm. Furthermore, we denote $\sigma$ the relative standard deviation percentage in multiple runs. The objective value, running time and performance ratio on average are denoted as $\gamma_{avg}, T_{avg}$, and $\rho_{avg}$. In addition, the geometric mean, which is more appropriate than the arithmetic mean when analysing normalized performance numbers, is used to calculate the values of $\rho_{avg}, T_{avg}$ (see \cite{fleming1986not} for more information).

CPLEX 12.6.2 is used whenever the MILP formulation needs to be solved, and optimal TSP tours are obtained with the state-of-the-art Concorde solver \cite{applegate2006concorde}. The values of $k$ in $k$-nearest neighbour and $k$-cheapest insertion heuristics are chosen randomly between $\{2,3\}$ to give the best results. Also by experiment, the value of parameter $n_{TSP}$ of \textit{GRASP} is set to 2000 in all tests. And finally, all instances and detailed results are available at \url{http://research.haquangminh.com/tspd/index}.

\section{Results}
\label{section:results}
In this section, we present and analyse the computational results obtained by the proposed methods. The algorithms are implemented in C++ and run on an Intel Core i7-6700 @ 3.4 GHz processor. Different experiments have been carried out to evaluate the performance of the proposed methods and analyse the impact of parameters: explore the performance of GRASP on different TSP-tour construction heuristics in min-cost TSP-D, compare min-cost TSP-D solutions provided by the proposed heuristics and optimal solutions computed from the MILP formulation (if possible), compare min-cost TSP-D solutions with TSP solutions (i.e., no drone delivery), compare GRASP with TSP-LS on min-cost TSP-D instances, analyse the impact of the drone/truck cost ratio in min-cost TSP-D, and verify heuristics' performance under min-time objective as well as the trade-off between two objectives.

\subsection{Performance of GRASP on different TSP-tour construction heuristics in the min-cost TSP-D}

In this subsection, we evaluate the performance of GRASP under three proposed TSP construction heuristics in the min-cost TSP-D. We also analyse the impact of the local search operators on the behaviour of GRASP. For each instance set labeled from B to G, we select 3 instances. Then each combination of instance and TSP construction heuristic will be run 10 times. With 18 instances, 3 heuristics, 2 local search settings (enable/disable), we have in total: $18 \times 3 \times 10 \times 2 = 1080$ tests. We use TSP optimal solutions (obtained by Concorde) as reference $referenceValue$ to calculate the performance ratios $\rho$.

The columns $\rho^{tsp}_{avg}$ represent the performance ratio on average of TSP solutions obtained by TSP tour generation heuristics. The columns $\rho^{withLS}_{avg}$, $\rho^{noLS}_{avg}$ respectively report the performance ratio on average of GRASP with and without local search. The results are presented in Table \ref{table: grasp-perf}. 

\begin{table}[H]
\begin{center}
\scalebox{0.58}{
\begin{tabular}{ | c | c | c | c | c | c | c | c | c | c | c | c | c | c | c | c |}
\hline
Instance & \multicolumn{5}{|c|}{$k$-nearest neighbour} & \multicolumn{5}{|c|}{$k$-cheapest insertion} & \multicolumn{5}{|c|}{random insertion}
\\ \hline
	& $\rho^{withLS}_{avg}$ & $\sigma$ & $T_{avg}$ & $\rho^{tsp}_{avg}$ & $\rho^{noLS}_{avg}$ & $\rho^{withLS}_{avg}$ & $\sigma$ & $T_{avg}$ & $\rho^{tsp}_{avg}$ & $\rho^{noLS}_{avg}$ & $\rho^{withLS}_{avg}$ & $\sigma$ & $T_{avg}$ & $\rho^{tsp}_{avg}$ & $\rho^{noLS}_{avg}$
\\ \hline
B1 & 66.33 & 1.26 & 8.57 & 142.12 & 82.55 & 66.80 & 0.27 & 6.94 & 117.31 & 77.20 & 69.92 & 1.44 & 66.70 & 409.23 & 187.15\\ \hline
B2 & 74.33 & 0.66 & 8.66 & 146.05 & 82.25 & 75.72 & 1.45 & 4.56 & 115.86 & 82.79 & 75.80 & 1.19 & 58.83 & 420.47 & 191.41\\ \hline
B3 & 71.62 & 1.22 & 9.90 & 137.05 & 85.11 & 75.17 & 0.10 & 6.09 & 117.38 & 86.03 & 73.80 & 1.49 & 57.38 & 438.95 & 209.62\\ \hline
C1 & 71.11 & 1.01 & 7.09 & 143.08 & 82.05 & 76.66 & 1.42 & 6.23 & 120.74 & 85.74 & 75.44 & 1.37 & 43.12 & 462.91 & 345.88\\ \hline
C2 & 72.66 & 0.91 & 8.69 & 150.82 & 81.78 & 78.72 & 0.83 & 6.54 & 122.71 & 84.65 & 75.80 & 0.74 & 52.65 & 498.46 & 363.96\\ \hline
C3 & 81.09 & 1.58 & 5.44 & 147.67 & 89.66 & 83.52 & 0.41 & 4.67 & 115.72 & 87.64 & 82.46 & 1.30 & 43.32 & 521.19 & 389.58\\ \hline
D1 & 77.63 & 0.79 & 6.97 & 146.24 & 92.73 & 81.16 & 0.30 & 4.85 & 118.39 & 89.14 & 79.36 & 1.30 & 58.33 & 469.68 & 364.12\\ \hline
D2 & 72.69 & 0.81 & 6.32 & 140.80 & 91.17 & 72.73 & 0.84 & 6.08 & 115.21 & 81.73 & 75.11 & 1.06 & 51.61 & 459.79 & 355.26\\ \hline
D3 & 74.75 & 0.74 & 5.85 & 144.54 & 88.93 & 85.51 & 0.47 & 4.27 & 123.19 & 90.78 & 78.04 & 1.47 & 52.83 & 519.60 & 388.39\\ \hline
E1 & 70.40 & 1.21 & 85.40 & 137.31 & 84.81 & 69.69 & 0.65 & 37.35 & 107.67 & 76.71 & 78.07 & 2.12 & 605.80 & 554.25 & 286.11\\ \hline
E2 & 70.64 & 1.13 & 82.52 & 135.61 & 85.71 & 67.16 & 0.46 & 41.68 & 112.61 & 74.89 & 79.31 & 1.79 & 605.82 & 560.64 & 278.69\\ \hline
E3 & 71.29 & 0.64 & 85.59 & 135.23 & 85.79 & 70.70 & 0.60 & 37.46 & 108.42 & 76.58 & 78.93 & 1.91 & 605.99 & 561.45 & 283.99\\ \hline
F1 & 75.12 & 1.27 & 67.42 & 144.97 & 94.16 & 78.12 & 0.56 & 51.06 & 115.20 & 84.60 & 86.37 & 1.27 & 604.38 & 614.22 & 512.49\\ \hline
F2 & 74.62 & 1.40 & 81.67 & 148.48 & 93.44 & 77.08 & 1.19 & 73.21 & 120.45 & 86.02 & 82.57 & 2.44 & 604.33 & 647.65 & 529.60\\ \hline
F3 & 76.44 & 1.27 & 80.90 & 137.19 & 93.56 & 79.26 & 0.91 & 63.74 & 120.15 & 87.27 & 85.02 & 2.41 & 604.39 & 662.81 & 527.52\\ \hline
G1 & 78.18 & 1.81 & 69.73 & 149.72 & 95.99 & 81.37 & 0.87 & 64.65 & 125.09 & 89.75 & 87.29 & 1.51 & 604.16 & 682.44 & 570.93\\ \hline
G2 & 78.28 & 1.36 & 90.76 & 148.37 & 97.76 & 81.96 & 0.73 & 84.38 & 118.48 & 89.63 & 88.66 & 0.83 & 604.18 & 749.13 & 623.78\\ \hline
G3 & 74.19 & 1.04 & 89.86 & 146.62 & 95.86 & 72.22 & 0.99 & 84.66 & 120.03 & 82.26 & 81.75 & 2.18 & 604.20 & 662.59 & 556.53\\ \hline
Mean & \textbf{73.88} & & 24.44 & 143.35 & 88.92 & 76.12 & & 17.72 & 117.39 & 83.94 & 79.50 & & 179.69 & 541.23 & 362.85\\ 
\hline
\end{tabular}
}
\caption{Performance of GRASP on TSP-tour construction heuristics in min-cost TSP-D}
\label{table: grasp-perf}
\end{center}
\end{table}

In overall, GRASP with $k$-nearest neighbour provided the best performance in terms of solution quality, followed by $k$-cheapest insertion and then random insertion. It is well-known that greedy algorithms such as nearest neighbour and cheapest insertion give better solutions for the TSP than totally random insertion algorithm does (as confirmed again by the columns "$\rho^{tsp}_{avg}$"); the use of good TSP tours is an important factor to improve the quality of our GRASP. However, although $k$-cheapest insertion in general gives better TSP tours than $k$-nearest neighbour, TSP-D solutions obtained from $k$-nearest neighbour are better than ones obtained from $k$-cheapest insertion. The reason could be due to our local search operators which seems to work better with $k$-nearest neighbour. In GRASP with $k$-cheapest insertion, the local search operators in general converge more prematurely. And as a result, GRASP with $k$-cheapest insertion stops earlier than GRASP with $k$-nearest neighbour. In addition, we carried out some additional tests and found that, in general, using optimal TSP tours does not provide best solutions for the min-cost TSP-D.

Furthermore, all three heuristics provided stable results with most of standard deviations $\sigma$ less than 2~\%. More precisely, GRASPs with $k$-nearest neighbour and $k$-cheapest insertion are more stable than GRASP using random insertion heuristic. From these analyses, we decide to use $k$-nearest neighbour heuristic to generate TSP tours for GRASP in the next experiments.

\subsection{Comparison with min-cost TSP-D optimal solutions}

In this section, to validate the MILP formulation, we report the results obtained by CPLEX. We also wish to observe the possibility that finds optimal solutions of two approximate approaches GRASP and TSP-LS. The preliminary experiments show that the MILP formulation cannot solve to optimality instances with more than 10 nodes under a time limit of 1~hour. Therefore, in this subsection, we use only the 10-customer instances to compare the solutions obtained by GRASP, TSP-LS and ordinary TSP with the optimal solutions of the min-cost TSP-D computed through the MILP formulation. The $referenceValue$ used to compute the ratio $\rho$ is the optimal min-cost TSP-D solution. For each instance, GRASP is repeatedly run 10 times and we record the number of times (in Column $opt$) it can reach the optimality. The comparison results reported in Table \ref{table: compare-optimal} show that GRASP can find all optimal solutions consuming much less computation time than the MILP formulation. On the other hand, although TSP-LS is faster, it can only find one optimal solution. It is clear that GRASP outperforms TSP-LS in terms of solution quality. In details, GRASP shows a stable performance with standard deviation of 0 (which reported in Column $\sigma$) and can reach to optimality in all cases. From the column "TSP", we observe that using the drone allows to save more than 20~\% and up to 53~\% of operational costs. Next, we focus on analysing the performance of GRASP and TSP-LS on the larger instances. 

\begin{table}[H]
\begin{center}
\scalebox{0.7}{
\begin{tabular}{ | c | c | c | c | c | c | c | c | c | c | c | c | c |}
\hline
Instance & \multicolumn{2}{|c|}{TSP} & \multicolumn{2}{|c|}{MILP formulation} & \multicolumn{5}{|c|}{GRASP} & \multicolumn{3}{|c|}{TSP-LS} \\ \cline{2-13}
& $\gamma$ & $\rho$ & $\gamma$ & $T$ & $\gamma_{avg}$ & $T_{avg}$ & $\rho_{avg}$ & $\sigma$ & $opt$ & $\gamma$ & $T$ & $\rho$
\\ \hline
A1 & 1007.33 & 153.01 & 658.322 & 46.64 & 658.322 & 0.84 & 100 & 0 & 10 & 810.244 & 0.013 & 123.07 \\ 
A2 & 955.876 & 140.58 & 679.932 & 144.51 & 679.932 & 1.06 & 100 & 0 & 10 & 777.119 & 0.006 & 114.29 \\ 
A3 & 985.679 & 120.31 & 819.251 & 133.30 & 819.251 & 0.78 & 100 & 0 & 10 & 819.251 & 0.005 & 100 \\ 
A4 & 944.645 & 126.22 & 748.405 & 41.31 & 748.405 & 1.70 & 100 & 0 & 10 & 834.89 & 0.007 & 111.55 \\ 
A5 & 985.679 & 121.60 & 810.567 & 57.18 & 810.567 & 1.63 & 100 & 0 & 10 & 853.728 & 0.005 & 105.32 \\ 
\hline
\end{tabular}
}
\caption{Comparison with the min-cost TSP-D optimal solution.}
\label{table: compare-optimal}
\end{center}
\end{table}

\subsection{Performance of heuristics on the larger instances in the min-cost TSP-D}
\label{section:performance-larger-instance}

In this subsection, we aim to analyse the performance of GRASP and TSP-LS on the larger min-cost TSP-D instances---those with 50 and 100 customers. Two methods TSP-LS and GRASP are tested and obtained solutions are compared with ones of the ordinary TSP. The $referenceValue$ used to compute the ratio $\rho$ is the objective value of the TSP optimal solution. For each instance, we also report the average waiting times of truck and drone as well as the average latest time at which either the truck or the drone return to the depot (Column $w_{avg}, w'_{avg}$ and $t_{avg}$). These values are measured in minutes. Again, for each instance, GRASP is repeatedly run 10 times. Tables \ref{table:grasp-comparison-50} and \ref{table:grasp-comparison-100} show the results for the instances with 50 and 100 customers, respectively. 

As can be observed, GRASP outperforms TSP-LS in terms of solution quality. In terms of running time, GRASP runs slower. However, considering that it never runs in more than 4 minutes, while performs up to 7~\% better than TSP-LS in terms of $\rho_{avg}$, this trade-off is worthy.

In all cases, GRASP finds the best solutions. Regardless of slower speed, its average computational time is acceptable on even 100-customer instances (about 2.5 minutes averagely). Furthermore, its relative standard deviation percentage -- reported in Column $\sigma$ -- is less than 3\% in all instances, proving the stability of the algorithm. The results obtained once again prove the effectiveness of using the drone for delivery. GRASP gives solutions with a cost saving of more than 25~\% compared with the TSP optimal solutions, which do not use any drone delivery.

Regarding the waiting times, one can observe that in min-cost TSP-D solutions, truck has to wait for drone most of the time among all instances ($w_{avg} > w'_{avg}$). In details, while drone's waiting times are only a couple of minutes, truck's waiting times make up approximately 25.95\% and 26.20\% of the delivery completion time ($t_{avg}$) regarding to geometric mean in 50-customer and 100-customer instances, respectively. This could be due to the fact that truck's transportation cost is much larger than drone's transportation cost (25 times larger), the min-cost TSP-D solutions tend to select drone deliveries in which flying distance of drone is quite longer than traveling distance of truck.


\begin{table}[H]
\begin{center}
\scalebox{0.85}{
\begin{tabular}{ | c | c | c | c | c | c | c | c | c | c | c |}
\hline
\textbf{$N$ = 50} & \multicolumn{8}{|c|}{GRASP} & \multicolumn{2}{|c|}{TSP-LS} \\ \cline{2-11}
& $\gamma_{best}$ & $\gamma_{avg}$ & $\rho_{avg}$ & $T_{avg}$ & $\sigma$ & $w_{avg}$ & $w'_{avg}$ & $t_{avg}$ & $\rho_{avg}$ & $T_{avg}$
\\ \hline

B1 & 1372.82 & 1413.24 & 66.59 & 16.30 & 1.34 & 70.48 & 0.56 & 192.52 & 78.62 & 0.40\\ \hline
B2 & 1491.30 & 1513.98 & 73.42 & 15.67 & 0.48 & 36.13 & 1.00 & 162.78 & 77.42 & 0.34\\ \hline
B3 & 1503.78 & 1521.67 & 72.06 & 16.70 & 0.68 & 44.11 & 1.57 & 165.46 & 81.44 & 0.28\\ \hline
B4 & 1396.17 & 1426.20 & 65.33 & 15.92 & 0.98 & 66.97 & 0.13 & 190.45 & 79.38 & 1.01\\ \hline
B5 & 1457.91 & 1500.90 & 71.52 & 18.73 & 1.51 & 53.38 & 1.94 & 178.12 & 81.28 & 0.39\\ \hline
B6 & 1316.08 & 1353.76 & 63.87 & 15.94 & 1.04 & 81.87 & 0.57 & 198.83 & 75.51 & 0.30\\ \hline
B7 & 1370.05 & 1399.71 & 65.90 & 14.27 & 0.83 & 63.16 & 1.46 & 183.53 & 78.69 & 0.30\\ \hline
B8 & 1484.93 & 1517.23 & 73.24 & 15.23 & 0.95 & 60.04 & 0.41 & 184.35 & 83.03 & 0.28\\ \hline
B9 & 1442.09 & 1468.86 & 70.27 & 17.05 & 0.94 & 43.65 & 3.87 & 168.32 & 79.19 & 0.31\\ \hline
B10 & 1392.54 & 1429.57 & 67.94 & 15.19 & 1.04 & 54.40 & 0.08 & 174.44 & 75.62 & 0.33\\ \hline
C1 & 2870.41 & 2935.87 & 71.70 & 12.62 & 0.88 & 112.76 & 2.47 & 318.15 & 79.52 & 0.12\\ \hline
C2 & 2804.47 & 2868.67 & 72.97 & 15.74 & 0.75 & 88.56 & 2.43 & 293.69 & 78.67 & 0.12\\ \hline
C3 & 3087.55 & 3185.09 & 81.87 & 9.73 & 1.07 & 56.35 & 4.03 & 272.77 & 83.06 & 0.14\\ \hline
C4 & 2844.10 & 2916.86 & 70.97 & 11.78 & 0.70 & 91.54 & 1.37 & 297.05 & 82.39 & 0.16\\ \hline
C5 & 3323.92 & 3367.34 & 80.54 & 11.40 & 0.57 & 58.89 & 4.72 & 286.26 & 89.21 & 0.09\\ \hline
C6 & 3433.99 & 3472.39 & 79.79 & 11.24 & 0.65 & 68.47 & 3.43 & 301.24 & 86.71 & 0.12\\ \hline
C7 & 3001.13 & 3047.71 & 71.92 & 12.86 & 0.64 & 105.33 & 0.58 & 317.15 & 80.75 & 0.11\\ \hline
C8 & 3481.17 & 3557.99 & 82.21 & 13.07 & 1.02 & 71.66 & 3.02 & 312.14 & 86.84 & 0.10\\ \hline
C9 & 3267.23 & 3306.38 & 75.35 & 11.56 & 0.40 & 85.48 & 0.46 & 311.73 & 80.09 & 0.27\\ \hline
C10 & 3291.20 & 3356.29 & 78.34 & 13.77 & 0.84 & 74.47 & 1.29 & 304.23 & 82.22 & 0.14\\ \hline
D1 & 4159.39 & 4389.24 & 76.86 & 12.87 & 1.61 & 93.40 & 1.61 & 382.41 & 89.35 & 0.10\\ \hline
D2 & 4275.46 & 4334.40 & 72.32 & 11.67 & 0.52 & 106.81 & 2.42 & 392.96 & 76.75 & 0.06\\ \hline
D3 & 4085.71 & 4191.08 & 75.25 & 11.06 & 1.01 & 92.96 & 1.59 & 368.91 & 82.21 & 0.07\\ \hline
D4 & 4612.46 & 4714.62 & 77.14 & 12.74 & 0.80 & 91.33 & 1.96 & 399.14 & 80.52 & 0.11\\ \hline
D5 & 4717.67 & 4793.39 & 80.26 & 11.70 & 0.79 & 77.54 & 0.97 & 390.97 & 82.89 & 0.06\\ \hline
D6 & 4405.02 & 4485.87 & 78.64 & 11.73 & 0.79 & 78.61 & 2.87 & 373.53 & 85.93 & 0.06\\ \hline
D7 & 4749.57 & 4796.23 & 82.77 & 15.06 & 0.46 & 68.48 & 7.74 & 384.65 & 86.69 & 0.07\\ \hline
D8 & 4143.03 & 4287.87 & 77.71 & 11.99 & 1.56 & 90.75 & 1.05 & 374.24 & 87.62 & 0.06\\ \hline
D9 & 4653.73 & 4688.16 & 76.11 & 13.39 & 0.50 & 86.84 & 4.82 & 392.72 & 86.34 & 0.17\\ \hline
D10 & 4260.60 & 4301.83 & 75.33 & 11.96 & 0.41 & 91.95 & 5.98 & 375.94 & 78.89 & 0.08\\ \hline
Mean &  &  & \textbf{74.09} & 13.46 & &  &  &  & 81.80 & 0.15\\ \hline
\end{tabular}
}
\caption{Performance of heuristics on 50-customer instances in the min-cost TSP-D}
\label{table:grasp-comparison-50}
\end{center}
\end{table}

\begin{table}[H]
\begin{center}
\scalebox{0.85}{
\begin{tabular}{ | c | c | c | c | c | c | c | c | c | c | c |}
\hline
\textbf{$N$ = 100} & \multicolumn{8}{|c|}{GRASP} & \multicolumn{2}{|c|}{TSP-LS} \\ \cline{2-11}
& $\gamma_{best}$ & $\gamma_{avg}$ & $\rho_{avg}$ & $T_{avg}$ & $\sigma$ & $w_{avg}$ & $w'_{avg}$ & $t_{avg}$ & $\rho_{avg}$ & $T_{avg}$
\\ \hline

E1 & 2206.53 & 2255.99 & 70.64 & 137.02 & 0.96 & 81.22 & 1.75 & 293.35 & 76.22 & 5.52\\ \hline
E2 & 2210.61 & 2273.09 & 70.53 & 136.68 & 1.08 & 77.09 & 2.21 & 288.01 & 76.26 & 5.90\\ \hline
E3 & 2248.16 & 2312.76 & 71.43 & 148.62 & 1.09 & 77.57 & 1.71 & 287.39 & 72.41 & 6.09\\ \hline
E4 & 2179.06 & 2223.97 & 70.35 & 178.37 & 0.90 & 75.97 & 1.69 & 282.29 & 78.14 & 6.02\\ \hline
E5 & 2286.16 & 2360.30 & 73.58 & 172.10 & 1.31 & 60.15 & 2.26 & 269.52 & 77.85 & 6.17\\ \hline
E6 & 2244.62 & 2313.86 & 71.89 & 195.51 & 1.05 & 74.50 & 1.87 & 286.99 & 78.14 & 5.89\\ \hline
E7 & 2249.09 & 2313.67 & 71.94 & 190.84 & 0.90 & 67.23 & 2.28 & 279.10 & 82.33 & 6.32\\ \hline
E8 & 2220.88 & 2272.55 & 70.66 & 189.24 & 0.79 & 71.45 & 1.40 & 280.26 & 72.37 & 6.64\\ \hline
E9 & 2279.91 & 2326.29 & 72.33 & 172.03 & 0.83 & 67.60 & 1.72 & 277.53 & 74.74 & 5.72\\ \hline
E10 & 2324.74 & 2384.52 & 74.30 & 204.74 & 0.96 & 64.16 & 1.94 & 277.90 & 77.23 & 4.98\\ \hline
F1 & 4569.83 & 4648.20 & 76.20 & 111.07 & 0.85 & 109.53 & 6.36 & 443.43 & 83.13 & 1.23\\ \hline
F2 & 4186.76 & 4318.78 & 74.74 & 143.07 & 1.47 & 138.73 & 2.39 & 459.57 & 80.43 & 1.19\\ \hline
F3 & 4414.38 & 4563.64 & 76.57 & 146.75 & 1.31 & 119.39 & 6.88 & 454.68 & 81.77 & 1.46\\ \hline
F4 & 4499.09 & 4600.27 & 79.53 & 128.53 & 1.25 & 123.85 & 3.15 & 456.31 & 80.99 & 1.38\\ \hline
F5 & 4381.37 & 4597.32 & 76.34 & 159.76 & 1.66 & 129.97 & 1.88 & 464.85 & 80.65 & 1.13\\ \hline
F6 & 4032.90 & 4171.80 & 74.54 & 157.70 & 1.53 & 130.63 & 3.55 & 442.99 & 79.74 & 1.06\\ \hline
F7 & 4076.31 & 4213.52 & 72.64 & 170.14 & 1.62 & 159.33 & 1.44 & 478.18 & 74.39 & 1.29\\ \hline
F8 & 4491.20 & 4597.90 & 75.37 & 165.96 & 0.98 & 126.90 & 4.52 & 464.09 & 82.89 & 1.31\\ \hline
F9 & 4388.91 & 4463.39 & 75.24 & 153.04 & 0.90 & 124.91 & 3.43 & 455.09 & 83.62 & 1.52\\ \hline
F10 & 4173.64 & 4567.84 & 76.48 & 153.99 & 2.31 & 118.52 & 3.07 & 451.57 & 80.48 & 1.51\\ \hline
G1 & 5947.97 & 6148.50 & 77.05 & 116.24 & 1.61 & 163.73 & 3.72 & 589.97 & 79.66 & 0.65\\ \hline
G2 & 5882.97 & 5987.64 & 79.70 & 158.00 & 0.76 & 118.74 & 5.14 & 532.92 & 81.70 & 0.52\\ \hline
G3 & 6074.57 & 6138.94 & 74.64 & 169.26 & 0.80 & 163.40 & 2.78 & 585.57 & 78.02 & 1.03\\ \hline
G4 & 6458.96 & 6632.14 & 82.34 & 143.47 & 1.16 & 135.84 & 5.02 & 588.44 & 85.79 & 0.90\\ \hline
G5 & 6198.95 & 6329.25 & 80.46 & 155.52 & 0.73 & 127.68 & 4.00 & 563.97 & 82.04 & 0.58\\ \hline
G6 & 6049.34 & 6343.26 & 77.02 & 177.07 & 1.69 & 149.52 & 6.42 & 589.69 & 81.67 & 0.64\\ \hline
G7 & 5889.08 & 6023.11 & 75.66 & 171.24 & 0.88 & 141.96 & 4.50 & 557.86 & 75.98 & 0.81\\ \hline
G8 & 5599.55 & 5871.96 & 71.99 & 156.90 & 1.87 & 159.24 & 6.95 & 570.88 & 80.03 & 0.86\\ \hline
G9 & 6050.80 & 6254.50 & 74.29 & 184.69 & 1.48 & 174.62 & 2.73 & 609.20 & 80.87 & 0.89\\ \hline
G10 & 6249.69 & 6534.13 & 79.63 & 162.11 & 1.97 & 124.85 & 6.79 & 572.85 & 83.47 & 0.78\\ \hline
Mean & & & \textbf{74.87} & 158.77 & & & & & 79.36 & 1.79\\ \hline
\end{tabular}
}
\caption{Performance of heuristics on 100-customer instances in the min-cost TSP-D}
\label{table:grasp-comparison-100}
\end{center}
\end{table}

\subsection{Impact of cost ratio in the min-cost TSP-D}

In this experiment, we explore the impact of the drone/truck cost ratio on the objective values of the min-cost TSP-D solutions provided by the GRASP and TSP-LS algorithms. By default, this parameter is set to 1:25; therefore, we added two more ratios, 1:10 and 1:50. Table \ref{table:cost-ratio} shows the geometric mean values of $\rho_{avg}$ for the two heuristics. The $referenceValue$ used to compute the ratio $\rho$ is the objective value of the TSP optimal solution. Again, for each instance, GRASP is repeatedly run 10 times.

Logically, the value of $\rho_{avg}$ should decrease as the ratio increases. However, it does not reduce proportionally. More specifically, for GRASP, when the ratio changes from 1:10 to 1:25, the mean of $\rho_{avg}$ decreases by approximately 5\% for the 50-customer instances and approximately 6\% for the 100-customer instances. In contrast, as the ratio changes from 1:25 to 1:50, the mean of $\rho_{avg}$ decreases by only approximately 3\% in both cases. The same phenomenon is observed for TSP-LS. Consequently, when constructing distribution networks for drone/truck combinations, overestimating the transportation cost of the drone does not always improve significantly the results. This means that the efficiency of investment in improving the cost ratio should be carefully considered because such an investment may prove more expensive than the savings in operational costs.

Varying the cost ratio does not significantly impact the relative performance between the heuristics. The GRASP still outperforms the TSP-LS in all cases in terms of solution quality but is slower in terms of running time.

\begin{small}
\begin{table}[H]
\begin{center}
\scalebox{0.8}{
\begin{tabular}{ | l | c | c | c | c |} 
\hline
\textbf{} & \multicolumn{2}{|c|}{$N$ = 50} & \multicolumn{2}{|c|}{$N$=100} \\ 
\cline{2-5}
Cost ratio & GRASP & TSP-LS & GRASP & TSP-LS \\ 
\hline
1:10 & 79.41 & 84.94 & 80.63 & 82.92 \\
1:25 & 74.09 & 81.80 & 74.86 & 79.36 \\
1:50 & 71.93 & 80.70 & 72.53 & 78.05 \\
\hline
\end{tabular}
}
\caption{Performance of heuristics with different cost-ratio settings in min-cost TSP-D.}
\label{table:cost-ratio}
\end{center}
\end{table}
\end{small}

\subsection{Performance of heuristics with min-time objective}

In this section, we analyse the performance of proposed algorithms under min-time objective. We first compare the solutions provided by GRASP with the best ones found by \cite{murray2015flying} on 10-customer instances. We then evaluate the performance of these two heuristics on the larger instances proposed in Section \ref{section:experiments}.

\subsubsection{Performance on small instances}

We now compare the performance of GRASP with Murray et al. - the best recorded results found in \cite{murray2015flying} on small size instances of 10 customers spreading in a region of 8-mile square. These results have been selected from different approaches proposed in \cite{murray2015flying} including MILP formulation with Gurobi solver and FSTSP heuristic with different TSP tour constructions (IP, Savings, Nearest Neighbor, and Sweep). The detailed results are presented in Table \ref{table:compare-fstsp}. The best found solutions are appeared in bold. Column $\epsilon$ represents the drone's endurance in minutes, while column "TSP" contains the optimal TSP solution values.

In overall, GRASP performs better than the methods presented in \cite{murray2015flying}. Among 72 instances, GRASP provides solutions worse than those of Murray et al. in only three instances while improves the results of Murray et al. in 20 instances. These results demonstrate the performance of our algorithm to solve not only min-cost TSP-D but also min-time TSP-D.

\begin{table}[H]
\centering
\scalebox{0.75}{
\begin{tabular}{|c|c|c|c|c|c|c|c|c|c|}
\hline
Instance           & $\epsilon$ & TSP    & Murray et al.  & GRASP           & Instance           & $\epsilon$ & TSP    & Murray et al.  & GRASP           \\ \hline
37v1  & 20        & 57.446 & \textbf{56.468} & \textbf{56.468} & 40v7  & 20        & 60.455 & 49.996          & \textbf{49.470} \\
37v1  & 40        & 57.446 & \textbf{50.573} & \textbf{50.573} & 40v7  & 40        & 60.455 & \textbf{49.204} & 49.233          \\
37v2  & 20        & 54.184 & \textbf{53.207} & \textbf{53.207} & 40v8  & 20        & 73.255 & 62.796          & \textbf{62.270} \\
37v2  & 40        & 54.184 & \textbf{47.311} & \textbf{47.311} & 40v8  & 40        & 73.255 & 62.270          & \textbf{62.033} \\
37v3  & 20        & 54.664 & \textbf{53.687} & \textbf{53.687} & 40v9  & 20        & 54.517 & 42.799          & \textbf{42.533} \\
37v3  & 40        & 54.664 & \textbf{53.687} & \textbf{53.687} & 40v9  & 40        & 54.517 & 42.799          & \textbf{42.533} \\
37v4  & 20        & 67.464 & \textbf{67.464} & \textbf{67.464} & 40v10 & 20        & 54.055 & \textbf{43.076} & \textbf{43.076} \\
37v4  & 40        & 67.464 & \textbf{66.487} & \textbf{66.487} & 40v10 & 40        & 54.055 & \textbf{43.076} & \textbf{43.076} \\
37v5  & 20        & 58.022 & 50.551          & \textbf{47.457} & 40v11 & 20        & 60.455 & \textbf{49.204} & \textbf{49.204} \\
37v5  & 40        & 58.022 & \textbf{45.835} & \textbf{45.835} & 40v11 & 40        & 60.455 & \textbf{49.204} & \textbf{49.204} \\
37v6  & 20        & 54.184 & 45.176          & \textbf{45.145} & 40v12 & 20        & 73.255 & \textbf{62.004} & \textbf{62.004} \\
37v6  & 40        & 54.184 & 45.863          & \textbf{44.602} & 40v12 & 40        & 73.255 & \textbf{62.004} & \textbf{62.004} \\
37v7  & 20        & 54.664 & \textbf{49.581} & \textbf{49.581} & 43v1  & 20        & 69.586 & \textbf{69.586} & \textbf{69.586} \\
37v7  & 40        & 54.664 & \textbf{46.621} & 46.754          & 43v1  & 40        & 69.586 & \textbf{55.493} & \textbf{55.493} \\
37v8  & 20        & 67.464 & \textbf{62.381} & \textbf{62.381} & 43v2  & 20        & 72.146 & \textbf{72.146} & \textbf{72.146} \\
37v8  & 40        & 67.464 & 59.776          & \textbf{59.614} & 43v2  & 40        & 72.146 & \textbf{58.053} & \textbf{58.053} \\
37v9  & 20        & 58.022 & 45.985          & \textbf{42.585} & 43v3  & 20        & 77.344 & \textbf{77.344} & \textbf{77.344} \\
37v9  & 40        & 58.022 & \textbf{42.416} & \textbf{42.416} & 43v3  & 40        & 77.344 & \textbf{69.175} & \textbf{69.175} \\
37v10 & 20        & 54.184 & 42.416          & \textbf{41.908} & 43v4  & 20        & 90.144 & \textbf{90.144} & \textbf{90.144} \\
37v10 & 40        & 54.184 & 41.729          & \textbf{40.908} & 43v4  & 40        & 90.144 & \textbf{82.700} & \textbf{82.700} \\
37v11 & 20        & 54.664 & \textbf{42.896} & \textbf{42.896} & 43v5  & 20        & 69.586 & 55.493          & \textbf{53.053} \\
37v11 & 40        & 54.664 & \textbf{42.896} & \textbf{42.896} & 43v5  & 40        & 69.586 & 53.447          & \textbf{52.093} \\
37v12 & 20        & 67.464 & \textbf{56.696} & \textbf{56.696} & 43v6  & 20        & 72.146 & 58.053          & \textbf{55.209} \\
37v12 & 40        & 67.464 & \textbf{55.696} & \textbf{55.696} & 43v6  & 40        & 72.146 & \textbf{52.329} & \textbf{52.329} \\
40v1  & 20        & 54.517 & \textbf{49.430} & \textbf{49.430} & 43v7  & 20        & 77.344 & \textbf{64.409} & \textbf{64.409} \\
40v1  & 40        & 54.517 & \textbf{46.886} & \textbf{46.886} & 43v7  & 40        & 77.344 & \textbf{60.743} & 60.886          \\
40v2  & 20        & 54.055 & \textbf{50.708} & \textbf{50.708} & 43v8  & 20        & 90.144 & \textbf{77.209} & \textbf{77.209} \\
40v2  & 40        & 54.055 & \textbf{46.423} & \textbf{46.423} & 43v8  & 40        & 90.144 & 73.967          & \textbf{73.727} \\
40v3  & 20        & 60.455 & \textbf{56.102} & \textbf{56.102} & 43v9  & 20        & 69.586 & 49.049          & \textbf{46.931} \\
40v3  & 40        & 60.455 & \textbf{53.933} & \textbf{53.933} & 43v9  & 40        & 69.586 & 47.250          & \textbf{46.931} \\
40v4  & 20        & 73.255 & \textbf{69.902} & \textbf{69.902} & 43v10 & 20        & 72.146 & \textbf{47.935} & \textbf{47.935} \\
40v4  & 40        & 73.255 & 68.397          & \textbf{67.917} & 43v10 & 40        & 72.146 & \textbf{47.935} & \textbf{47.935} \\
40v5  & 20        & 54.517 & \textbf{43.533} & \textbf{43.533} & 43v11 & 20        & 77.344 & 57.382          & \textbf{56.395} \\
40v5  & 40        & 54.517 & \textbf{43.533} & \textbf{43.533} & 43v11 & 40        & 77.344 & \textbf{56.395} & \textbf{56.395} \\
40v6  & 20        & 54.055 & \textbf{44.076} & \textbf{44.076} & 43v12 & 20        & 90.144 & \textbf{69.195} & \textbf{69.195} \\
40v6  & 40        & 54.055 & \textbf{44.076} & \textbf{44.076} & 43v12 & 40        & 90.144 & \textbf{69.195} & \textbf{69.195} 
\\ \hline
\end{tabular}
}

\caption{Comparison between GRASP and the best solutions found by \cite{murray2015flying}}
\label{table:compare-fstsp}
\end{table}

\

\subsubsection{Performance on larger instances}
\label{section:min-time-larger-instance}

Since larger instances of the min-time TSP-D are not available in the literature, we use the instances proposed in this work to analyse further the performance of GRASP and TSP-LS under min-time objective. Moreover, we vary the drone's speed among three values 25~km/h, 40~km/h and 55~km/h as in \cite{murray2015flying}. Again, GRASP is repeatedly run 10 times for each combination of instance and drone's speed. Some preliminary experiments show that, on many min-time TSP-D instances, solutions provided by GRASP were worse than ones provided by TSP-LS and even ordinary TSP. Upon further investigation, we found out that very few drone deliveries were used and min-time TSP solutions were not far from TSP solutions. As a result, TSP-LS which constructs optimal TSP tours and then improves them with local search operators could perform better than GRASP. In this experiment, we use an additional setting of GRASP that we call GRASP+: only one iteration with an optimal TSP tour is performed.

The results are shown in Table \ref{table:grasp-comparison-time}. The columns and their descriptions are similar to Tables \ref{table:grasp-comparison-50} and \ref{table:grasp-comparison-100} in Section \ref{section:performance-larger-instance} except that the $referenceValue$ used to compute the ratio $\rho$ is the objective value of the TSP optimal solution regarding to traveling time (instead of transportation cost in the previous section). In addition, the average objective value of GRASP+ is reported in minutes in Column $\gamma_{avg}$.

As can be observed, GRASP has better performance than TSP-LS on 50-customer instances, but performs slightly worse than TSP-LS on 100-customer instances regarding to solution quality. Its solutions are even worse than those of optimal TSP on the instances of class "E". This phenomenon could be due to high launch and retrieve times of drone (totally 2 minutes while the average traveling time of truck between two customers is only about 30 seconds - See Table \ref{table:instances}) leading to the low frequency of using drone in min-time TSP-D solutions. This also leads to an observation that, on the instances generated in wider regions, more drone deliveries are used and more savings in objective values of the min-time TSP-D are created. For example, the instances of type "E" have averagely 590,144 possibilities of drone deliveries and 100 customers, but only less than 6 drone deliveries are used (as showed in Table \ref{table:trade-off} of the next section). But on the instances of types "F" and "G", the number of average used drone deliveries increases to 16 and 19, respectively. GRASP is still much slower than TSP-LS due to its higher-level characteristic, but similarly to Section \ref{section:performance-larger-instance}, its running time is acceptable, slightly more than a minute in terms of geometric mean.

In overall, GRASP+ performs better than TSP-LS on all instance classes in terms of solution quality. Regarding the running time, GRASP+ also runs much faster with the average running time less than 0.1 second. These demonstrate the performance of our Split and local search operators. 

In terms of waiting times (Columns $w_{avg}$ and $w'_{avg}$), we can observe an opposite phenomenon with respect to Section \ref{section:performance-larger-instance}. Unlike min-cost TSP-D solutions where truck has to wait for drone most of the time ($w_{avg} > w'_{avg}$), min-time solutions tend to choose drone deliveries in which drone has to wait for truck ($w_{avg} < w'_{avg}$). 

\begin{table}[H]
\begin{center}
\scalebox{0.75}{
\begin{tabular}{ | c | c | c | c | c | c | c | c | c | c |}
\hline
Instances & \multicolumn{5}{|c|}{GRASP+} & \multicolumn{2}{|c|}{GRASP} & \multicolumn{2}{|c|}{TSP-LS} \\ \cline{2-10}
& $\gamma_{avg}$ & $\rho_{avg}$ & $T_{avg}$ & $w_{avg}$ & $w'_{avg}$ & $\rho_{avg}$ & $T_{avg}$ & $\rho_{avg}$ & $T_{avg}$
\\ \hline

B & 121.2 & 95.88 & 0.01 & 2.51 & 21.64 & 96.16 & 25.64 & 97.56 & 0.22\\ \hline
C & 232.8 & 92.80 & 0.01 & 4.93 & 28.08 & 93.40 & 21.09 & 94.28 & 0.10\\ \hline
D & 323.4 & 92.49 & 0.01 & 9.14 & 29.14 & 92.84 & 22.33 & 94.08 & 0.08\\ \hline
E & 190.2 & 98.86 & 0.04 & 0.14 & 30.77 & 101.79 & 275.08 & 98.88 & 1.62\\ \hline
F & 334.8 & 94.49 & 0.03 & 2.97 & 52.84 & 97.86 & 151.60 & 96.88 & 0.44\\ \hline
G & 449.4 & 92.90 & 0.02 & 6.34 & 51.15 & 96.91 & 142.16 & 95.13 & 0.33\\ \hline
Mean & & \textbf{94.54} & 0.01 & & & 96.45 & 64.44 & 96.12 & 0.27\\ \hline

\end{tabular}
}
\caption{Performance of heuristics of min-time TSP-D}
\label{table:grasp-comparison-time}
\end{center}
\end{table}

\subsection{Comparison between min-cost and min-time objectives}

In this experiment, we compare the TSP-D under two objectives in terms of performance ratios and the frequency of using drones given the results from Sections \ref{section:performance-larger-instance} and \ref{section:min-time-larger-instance}. The experimental results are presented in Tables \ref{table:mintime-comparison} and \ref{table:trade-off}.

In Table \ref{table:mintime-comparison}, Columns $\rho_{min-time}^{time}$ and $\rho_{min-time}^{cost}$ represent the performance ratios of two objective values, i.e. delivery completion time and operational cost, of the best solutions found in min-time TSP-D problems. Similarly, Columns $\rho_{min-cost}^{time}$ and $\rho_{min-cost}^{cost}$ are delivery completion time and operational cost of the best solution found in min-cost TSP-D problems. Again, the $value$ in performance ratio is the optimal TSP solution calculated in time and cost, depending on the objective type. In overall, min-time TSP-D solutions not only reduce the delivery completion time but also the operational cost compared to corresponding optimal TSP solutions. In contrast, min-cost TSP-D solutions increase the completion time compared to optimal TSP solutions by up to 56.25~\% and 43.55~\% for 50 and 100-customer instances. However, min-cost TSP-D solutions can save averagely 30~\% of the operational cost compared to 20~\% of min-time TSP-D solutions. This proves the importance of the new min-cost objective function in the class of transportation problems with drone.

\begin{table}[H]
\begin{center}
\scalebox{0.85}{
\begin{tabular}{ | c | c | c | c | c |}
\hline
Instances & $\rho_{min-time}^{time}$ & $\rho_{min-time}^{cost}$ & $\rho_{min-cost}^{time}$ & $\rho_{min-cost}^{cost}$
\\ \hline
B & 91.83 & 80.60 & 156.25 & 62.10\\ \hline
C & 86.37 & 76.89 & 115.35 & 69.20\\ \hline
D & 81.82 & 77.29 & 108.31 & 71.34\\ \hline
E & 97.31 & 90.27 & 143.55 & 68.59\\ \hline
F & 90.88 & 81.31 & 128.44 & 69.88\\ \hline
G & 85.63 & 77.07 & 111.15 & 68.65\\ \hline
Mean & 88.83 & 80.44 & 125.99 & 68.23\\ 
\hline
\end{tabular}
}
\caption{Trade-off between min-cost and min-time in terms of performance ratio}
\label{table:mintime-comparison}
\end{center}
\end{table}

In Table \ref{table:trade-off}, we observe the number of times the drone is used in both min-cost TSP-D solutions (Column "Min-Cost") and min-time TSP-D solutions (Columns "Min-Time 25", "Min-Time 40" and "Min-Time 55" representing the cases where the drone speed is set respectively to 25~km/h, 40~km/h and 55~km/h). The results show that min-cost solutions tend to use the drone to service up to 40\% of the customers whereas min-time solutions only use the drone to service up to 22\% of the customers. This is because, as analysed above, using drone likely helps to reduce the transportation cost but this is not the case for the delivery completion time. Additionally, increasing drone speed leads to higher frequency of using drones.

\begin{table}[H]
\begin{center}
\scalebox{0.7}{
\begin{tabular}{ | c | c | c | c | c |}
\hline
Instances & \multicolumn{4}{|c|}{Number of drone uses} \\ \cline{2-5}
& Min-Cost & Min-Time 25 & Min-Time 40 & Min-Time 55
\\ \hline
B & 20.27 & 4.70 & 6.96 & 8.04\\ \hline
C & 18.19 & 4.43 & 9.46 & 11.27\\ \hline
D & 17.89 & 3.10 & 10.20 & 12.47\\ \hline
E & 39.95 & 5.30 & 5.77 & 5.83\\ \hline
F & 36.77 & 9.57 & 15.97 & 18.58\\ \hline
G & 36.44 & 8.73 & 18.66 & 21.90\\
\hline
\end{tabular}
}
\caption{The frequency of using drone in two objective functions}
\label{table:trade-off}
\end{center}
\end{table}

\section{Conclusion}
\label{section:conclusion}

This paper presents a new variant of the Traveling Salesman Problem with Drone (TSP-D) whose objective is to minimize the total operational costs including the transportation cost and the waiting penalties. We propose a model, a mixed integer linear programming formulation and two heuristic methods---GRASP and TSP-LS---to solve the problem. The MILP formulation is an extension of the mathematical model propose in \cite{murray2015flying}; TSP-LS is adapted from an existing heuristic, while GRASP is based on our new split algorithm and local search operators. Numerous experiments conducted on a variety of instances of both objective functions show the performance of our GRASP algorithm. Overall, it outperforms TSP-LS in terms of solution quality in an acceptable running time. The results also demonstrate the important role of the new objective function in the new class of vehicle routing problems with drone. Future researches could aim for proposing more efficient metaheuristics which based on our split procedure. The extension of the proposed methods to solve problems with multiple vehicles and multiple drones could be also an interesting research direction.

\section*{Acknowledgement}
This research is funded by Vietnam National Foundation for Science and Technology Development (NAFOSTED) under grant number FWO.102.2013.04.

The authors thank Professor Chase C. Murray for his kind support by sending us the instances and detailed results of the min-time TSP-D problem.
\section*{References}

\bibliography{bibliography}

\end{document}